\documentclass[10pt,letterpaper]{article}

\usepackage{cvpr}
\usepackage{times}
\usepackage{epsfig}
\usepackage{graphicx}
\usepackage{amsmath}
\usepackage{amssymb, comment, microtype, kotex}

\makeatletter
\@namedef{ver@everyshi.sty}{}
\makeatother
\usepackage{booktabs,tikz,xcolor,float}

\usetikzlibrary{arrows,snakes,backgrounds,patterns,matrix,shapes,fit,calc,shadows,plotmarks}

\tikzset{>=stealth',every on chain/.append style={join},
         every join/.style={->}}

\usepackage[pagebackref=true,breaklinks=true,letterpaper=true,colorlinks,bookmarks=false]{hyperref}

\cvprfinalcopy 

\makeatletter
\newcommand{\printfnsymbol}[1]{%
  \textsuperscript{\@fnsymbol{#1}}%
}
\makeatother

\begin{document}

\title{Neural Sign Language Translation based on Human Keypoint Estimation}

\author{Sang-Ki Ko\thanks{Equal contribution} \and Chang Jo Kim\printfnsymbol{1} \and Hyedong Jung \and Choongsang Cho\\
Korea Electronics Technology Institute\\
22 Dawangpangyo-ro 712 beon-gil, Seongnam-Si, Gyeonggi-do 13488, South Korea\\
{\tt\small \{narame7,wowchangjo,hudson.keti,ideafisher.cho\}@gmail.com}
}

\maketitle

\begin{abstract}
We propose a sign language translation system based on human keypoint estimation. It is well-known that many problems in the field of computer vision require a massive amount of dataset to train deep neural network models. The situation is even worse when it comes to the sign language translation problem as it is far more difficult to collect high-quality training data. In this paper, we introduce the KETI (short for Korea Electronics Technology Institute) sign language dataset which consists of 14,672 videos of high resolution and quality. Considering the fact that each country has a different and unique sign language, the KETI sign language dataset can be the starting line for further research on the Korean sign language translation.
Using the KETI sign language dataset, we develop a neural network model for translating sign videos into natural language sentences by utilizing the human keypoints extracted from a face, hands, and body parts. The obtained human keypoint vector is normalized by the mean and standard deviation of the keypoints and used as input to our translation model based on the sequence-to-sequence architecture. As a result, we show that our approach is robust even when the size of the training data is not sufficient. Our translation model achieves 93.28\% (55.28\%, respectively) translation accuracy on the validation set (test set, respectively) for 105 sentences that can be used in emergency situations. We compare several types of our neural sign translation models based on different attention mechanisms in terms of classical metrics for measuring the translation performance.
\end{abstract}

\section{Introduction}
The absence of the ability to hear sounds is a huge obstacle to smooth and natural communication for the hearing-impaired people in a predominantly hearing world. In many social situations, the hearing-impaired people necessarily need help from professional sign language interpreters to communicate with the hearing people even when they have to reveal their very private and sensitive information. Moreover, the hearing-impaired people are more vulnerable in various emergency situations due to the communication barriers due to the absence of the hearing ability. As a consequence, the hearing-impaired people easily become isolated and withdrawn from society.
This leads us to investigate the possibility of developing an artificial intelligence technology
that understands and communicates with the hearing-impaired people.

However, sign language recognition or translation is a very challenging problem since the task involves a interpretation between visual and linguistic information. The visual information consists of several parts such as body movement and facial expression of a signer~\cite{ForsterSHKZPN12,von2008significance}. To interpret the collection of the visual information as natural language sentences is also one of tough challenges to realize the 
sign language translation problem.

In order to process a sequence, there have been several interesting variants of recurrent neural networks (RNNs) proposed including long short-term memory (LSTM)~\cite{HochreiterS97} and gated recurrent units (GRUs)~\cite{ChovGBBSB14}. These architectures have been successfully employed to resolve many problems involving the process of sequential data such as machine translation and image captioning~\cite{dai2017contrastive,liu2017attention,SutskeverVL14,xu2015show}. 
Moreover, many researchers working on the field of image and video understanding have raised the level that seemed infeasible even a few years ago by learning their neural networks with a massive amount of training data. Recently, many neural network models based on convolutional neural network (CNNs) exhibited excellent performances in various visual tasks such as image classification~\cite{husqueeze,huang2017densely}, object detection~\cite{gao2017dynamic,redmon2016you}, semantic segmentation~\cite{long2015fully,zhang2018context}, and action recognition~\cite{donahue2015long,luvizon20182d}.  

Understanding sign languages requires a high level of spatial and temporal understanding and therefore, is regarded as very difficult with the current level of computer vision and machine learning technology~\cite{DongLY15,ForsterSHKZPN12,gattupalli2016evaluation,KishoreSK14,KollerFN15,koller2017re,StarnerP95}. It should be noted that sign languages are different from hand (finger) languages as the hand languages only represent each letter in an alphabet with the shape of a single hand~\cite{CamgozHKNB18} while the linguistic meaning of each sign is determined by subtle difference of shape and movement of body, hands, and sometimes by facial expression of the signer~\cite{von2008significance}. 
More importantly, the main difficulty comes from the lack of dataset for training neural networks. Many sign languages represent different words and sentences of spoken languages with temporal sequences of gestures comprising continuous pose of hands and facial expressions. This implies that there are uncountably many combinations of the cases even to describe a single human intention with the sign language. 

Hence, we restrict ourselves to the task of translating sign language in various emergency situations. We construct the first Korean sign language dataset collected from fourteen professional signers who are actually hearing-impaired people and named it the KETI sign language dataset. The KETI sign language dataset consists of 14,672 high-resolution videos that recorded the Korean signs corresponding to 419 words and 105 sentences related to various emergency situations.
Using the KETI sign language dataset, we present our sign language translation model based on the well-known off-the-shelf human keypoint detector and the sequence-to-sequence translation model. To the best of our knowledge, this paper is the first to exploit the human keypoints for the sign language translation problem. Due to the inherent complexity of the dataset and the problem, we present an effective normalization technique for the extracted human keypoints to be used in the sign language translation. We implement the proposed ideas and conduct various experiments to verify the performance of the ideas with the test dataset.

The main contributions of this paper are highlighted as follows:
\begin{enumerate}
    \item We introduce the first large-scale Korean sign language dataset for sign language translation.
    \item We propose a sign language translation system based on the 2D coordinates of human keypoints estimated from sign videos.
    \item We present an effective normalization technique for preprocessing the 2D coordinates of human keypoints.
    \item We verify the proposed idea by conducting various experiments with the sign language dataset.
\end{enumerate}

\section{Related Work}

There have been many approaches to recognize hand languages that are used to describe letters of the alphabet with a single hand. It is relatively easier than recognizing sign languages as each letter of the alphabet simply corresponds to a unique hand shape. In~\cite{DongLY15}, the authors have utilized depth cameras (Microsoft's Kinect) and the random forest algorithm to recognize the English alphabet and have shown 92\% recognition accuracy.
A pose estimation method of the upper body represented by seven key points was proposed for recognizing the American Sign Language (ASL)~\cite{gattupalli2016evaluation}. We also note that there has been an approach by Kim et al.~\cite{KimK16} to recognize the Korean hand language by analyzing latent features of hand images.

In general, researchers rely on the movements and shapes of both hands to recognize sign languages. Starner et al.~\cite{StarnerP95} have developed a real-time system based on Hidden Markov model (HMM) to recognize sentence-level ASL. They have demonstrated two experimental results: they have used solidly colored gloves to make tracking of hands easier in the first experiment and the second experiment have been conducted without gloves. They have claimed that the word accuracy of glove-based system is 99.2\% but the accuracy drops to 84.7\% if they do not use gloves. It should be noted that those accuracy can be reached because they have exploited the grammar to reviewing the errors of the recognition. The word accuracy without grammar and gloves is 74.5\%. 

On the other hand, there have been approach to automatically learning signs from weakly annotated data such as TV broadcasts by using subtitles provided simultaneously with the signs~\cite{BuehlerZE09,CooperB09,PfisterCZ13}. Following this direction, Forster et al. released the RWTH-PHOENIX-Weather 2012~\cite{ForsterSHKZPN12} and its extended version RWTH-PHOENIX-Weather 2014~\cite{ForsterSKBN14} that consist of weather forecasts recorded from German public TV and manually annotated using glosses and natural language sentences where time boundaries have been marked on the gloss and the sentence level. Based on the RWTH-PHOENIX-Weather corpus, Koller et al.~\cite{KollerFN15} have presented a statistical approach performing large vocabulary continuous sign language recognition across different signers. They have developed a continuous sign language recognition system that utilizes multiple information streams including the hand shape, orientation and position, the upper body pose, and face expression such as mouthing, eye brows and eye gaze. 

Until recently, there have been many attempts to recognize and translate sign language using deep learning (DL). Oberweger et al.~\cite{OberwegerWL15} have introduced and evaluated several architectures for CNNs to predict the 3D joint locations of a hand given a depth map. Kishore et al.~\cite{KishoreSK14} have developed a sign language recognition system that is robust in different video backgrounds by extracting signers using boundary and prior shape information. Then, the feature vector is constructed from the segmented signer and used as input to artificial neural network. An end-to-end sequence modelling using CNN-BiLSTM architecture usually used for gesture recognition was proposed for large vocabulary sign language recognition with RWTH-PHOENIX-Weather 2014~\cite{koller2017re}.

At the same time, one of the most interesting breakthroughs in neural machine translation or even in the entire DL was introduced under the name of `sequence-to-sequence (seq2seq)'~\cite{SutskeverVL14}. The seq2seq model relies on a common framework called an encoder-decoder model with RNN cells such as LSTMs or GRUs. The seq2seq model proved its effectiveness in many sequence generation tasks by achieving almost the human-level performance~\cite{SutskeverVL14}. Despite its effectiveness, the seq2seq model still has some drawbacks such as the input sequences of varying lengths being represented in fixed-size vectors and the vanishing gradient due to the long-term dependency between distant parts.

Camgoz et al.~\cite{CamgozHKNB18} formalized the sign language translation problem based on the pre-existing framework of neural machine translation with word and spatial embeddings for target sequences and sign videos, respectively.
They have proposed to utilize the seq2seq models to learn how to translate the spatio-temproal representation of signs into the spoken or written language.
Recently, researchers developed a simple sign language recognition system based on bidirectional GRUs which just classifies a given sign language video into one of the classes that are predetermined~\cite{KoSJ18}.

\section{KETI Sign Language Dataset}

\begin{figure}[t]
\centering
\includegraphics[height=1.4in, width=2.0in]{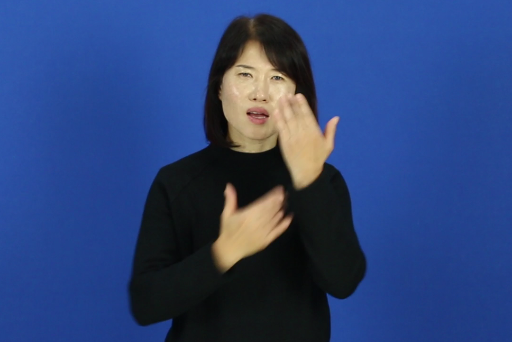}
\includegraphics[height=1.4in, width=2.0in]{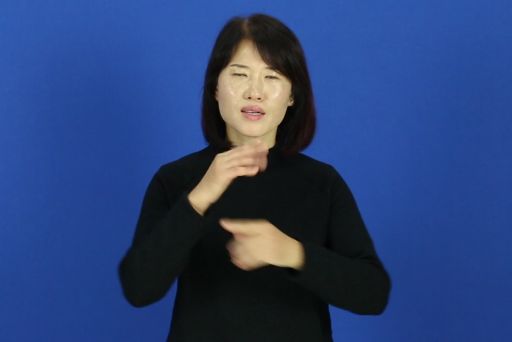}
\includegraphics[height=1.4in, width=2.0in]{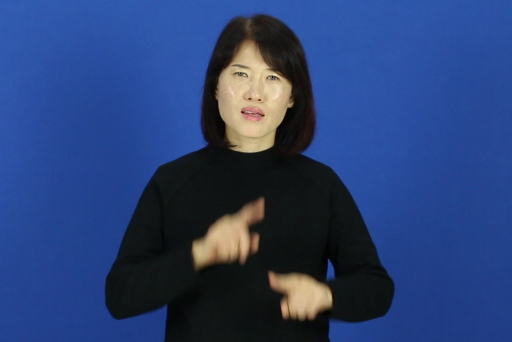}
\caption{Example frames from our sign language dataset. The frames are extracted in a temporal order from the video for the sentence `I am burned'.}
\label{fig:example_frame}
\end{figure}

\begin{figure}[t]
\centering
\includegraphics[height=1.69in, width=3.0in]{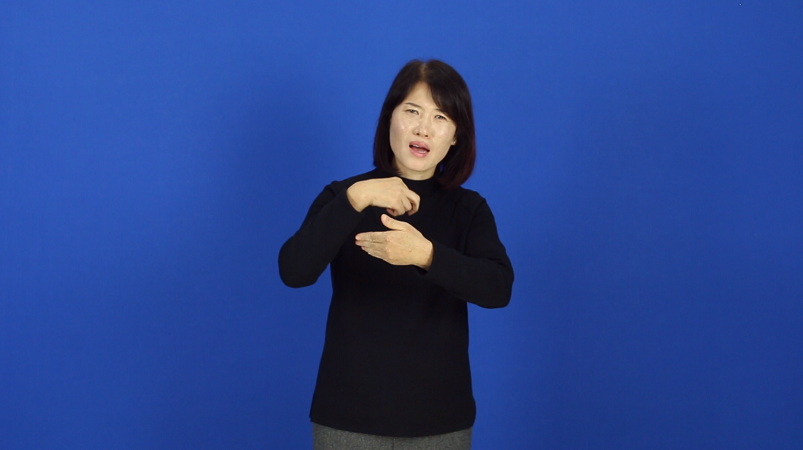}
\includegraphics[height=1.69in, width=3.0in]{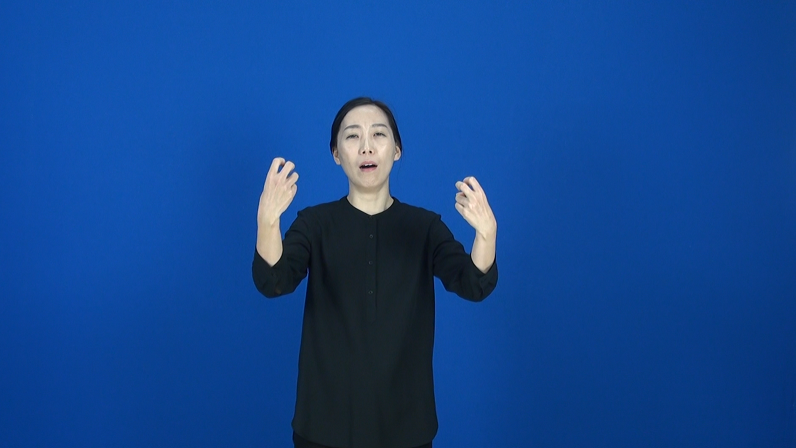}\\
\includegraphics[height=1.69in, width=3.0in]{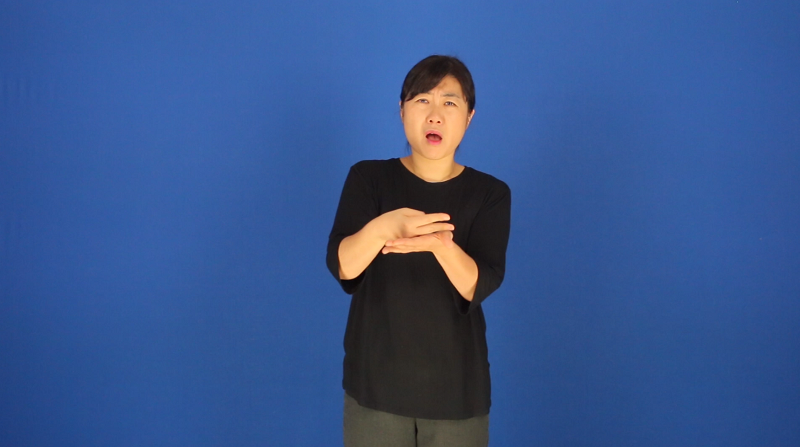}
\includegraphics[height=1.69in, width=3.0in]{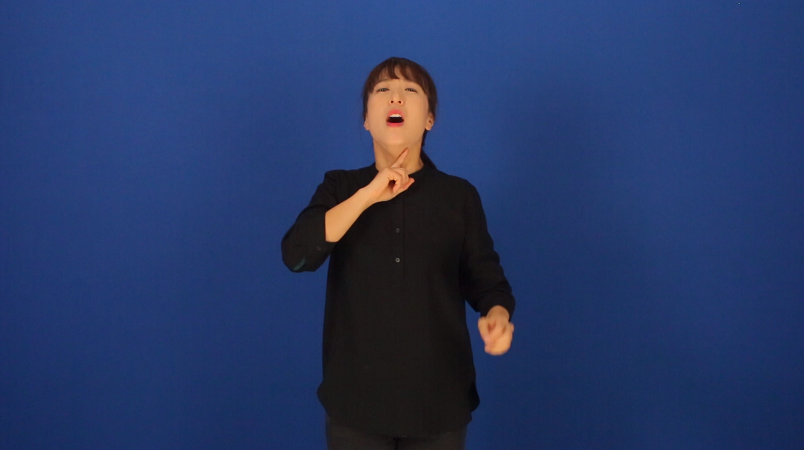}\\
\includegraphics[height=1.69in, width=3.0in]{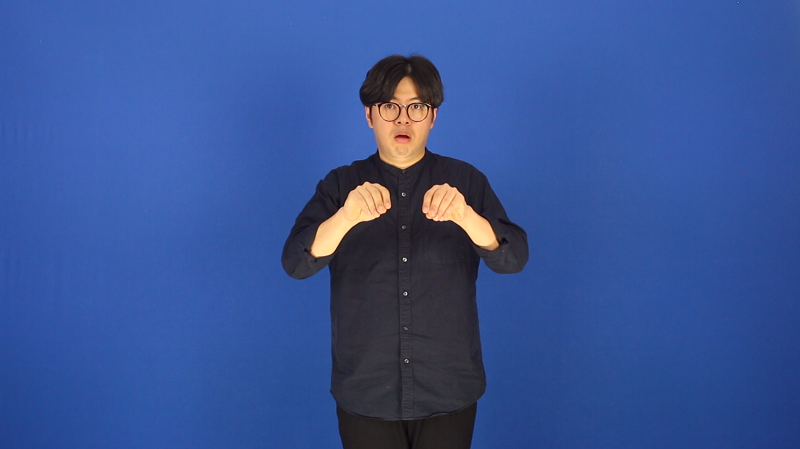}
\includegraphics[height=1.69in, width=3.0in]{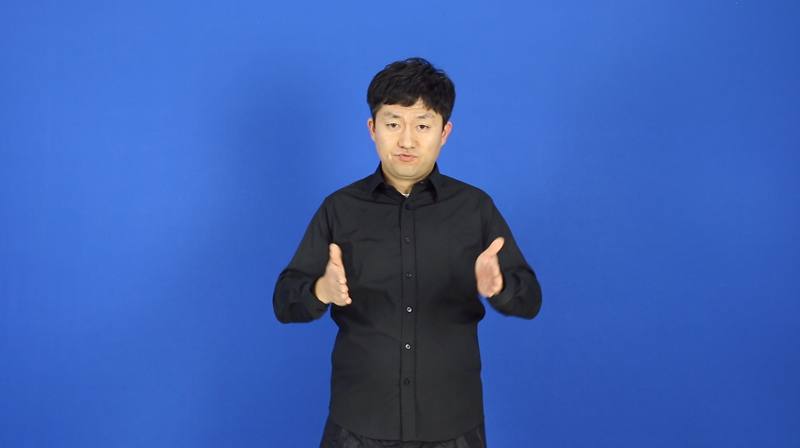}\\
\includegraphics[height=1.69in, width=3.0in]{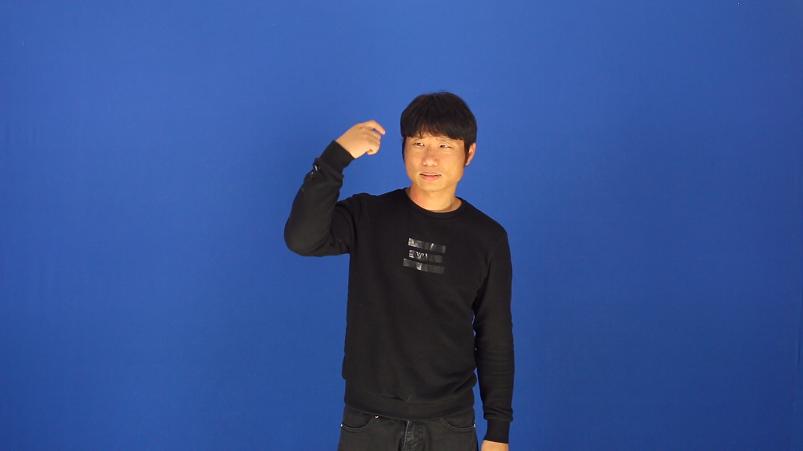}
\includegraphics[height=1.69in, width=3.0in]{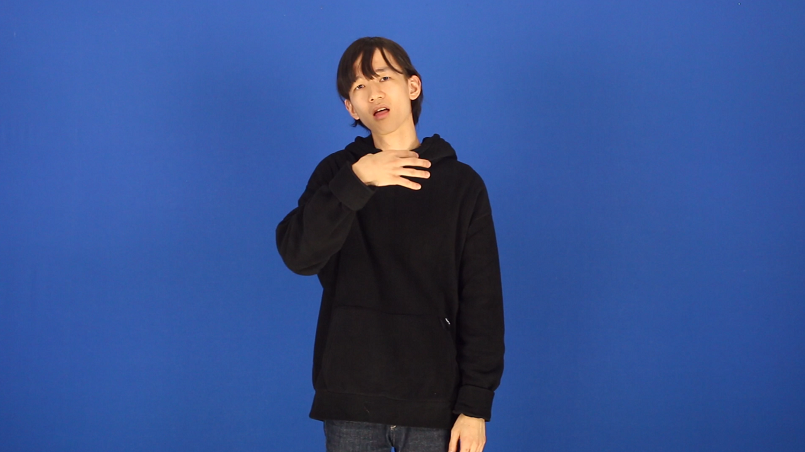}\\
\includegraphics[height=1.69in, width=3.0in]{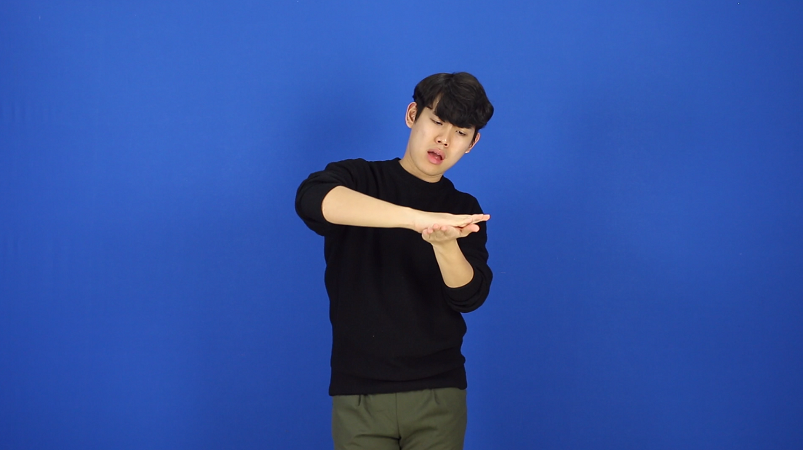}
\includegraphics[height=1.69in, width=3.0in]{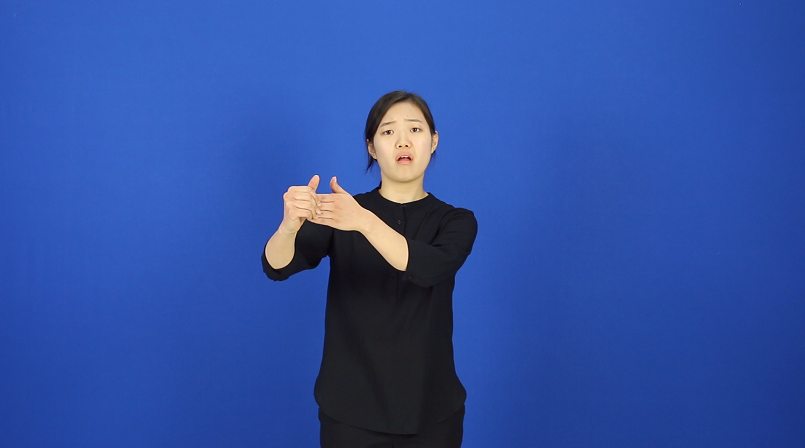}\\
\caption{Ten example frames from from our sign language dataset. Each frame is extracted from a sign video of a distinct signer.}
\label{fig:example_frame2}
\end{figure}

\begin{figure}[t]
\centering
\includegraphics[height=1.69in, width=3.0in]{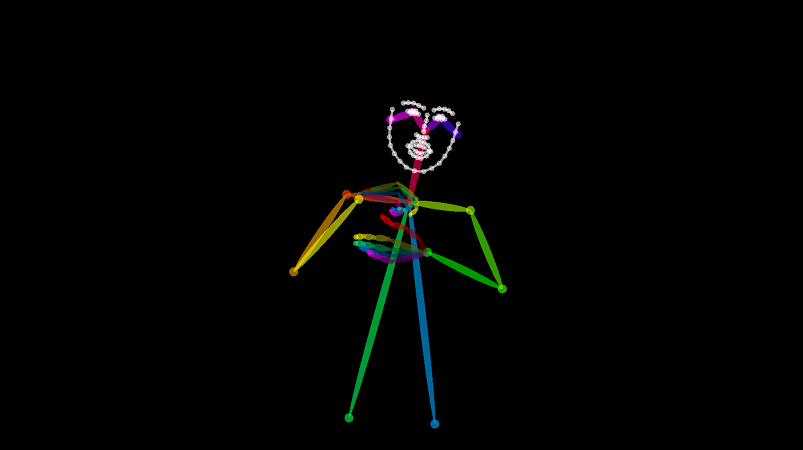}
\includegraphics[height=1.69in, width=3.0in]{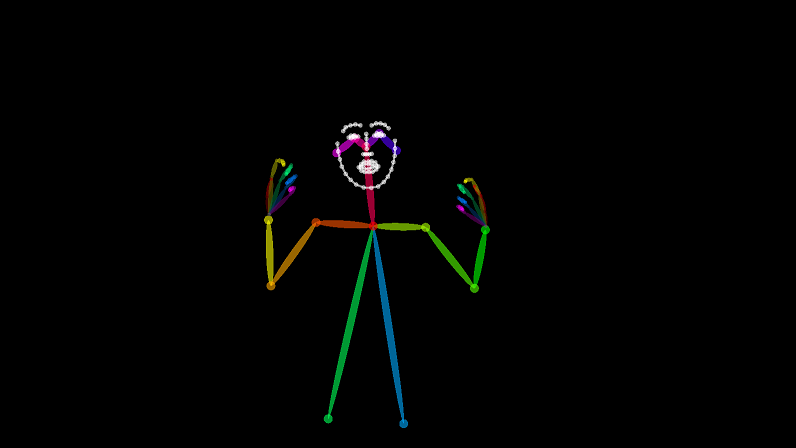}\\
\includegraphics[height=1.69in, width=3.0in]{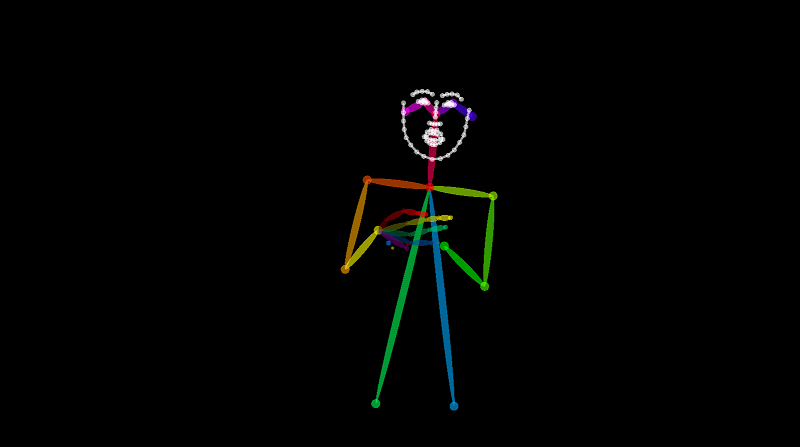}
\includegraphics[height=1.69in, width=3.0in]{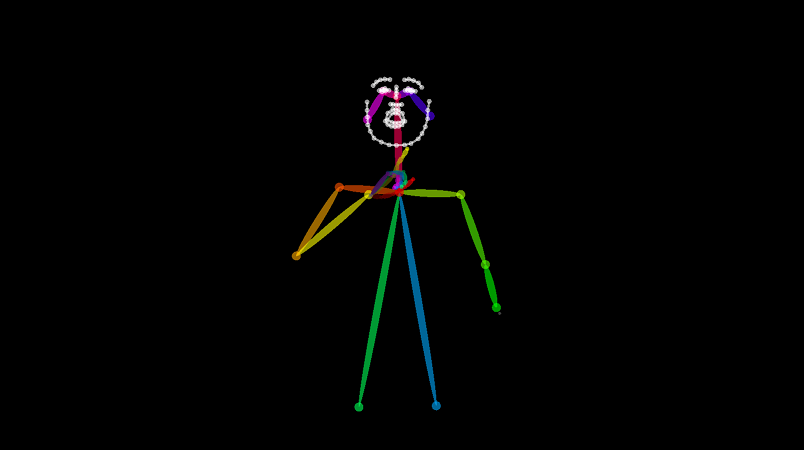}\\
\includegraphics[height=1.69in, width=3.0in]{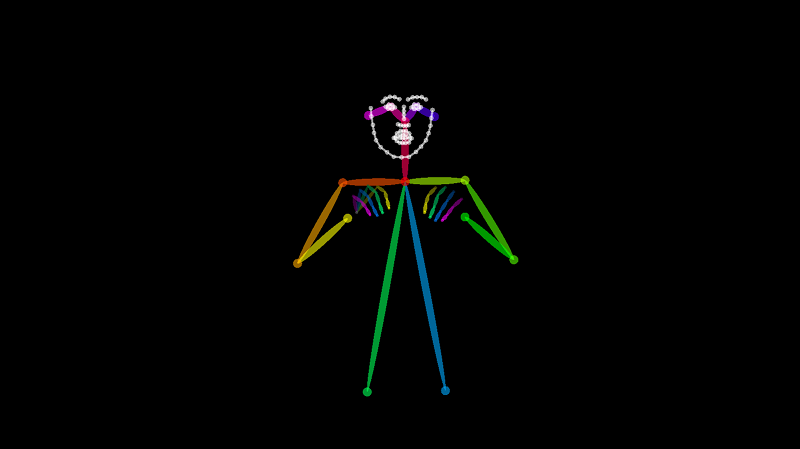}
\includegraphics[height=1.69in, width=3.0in]{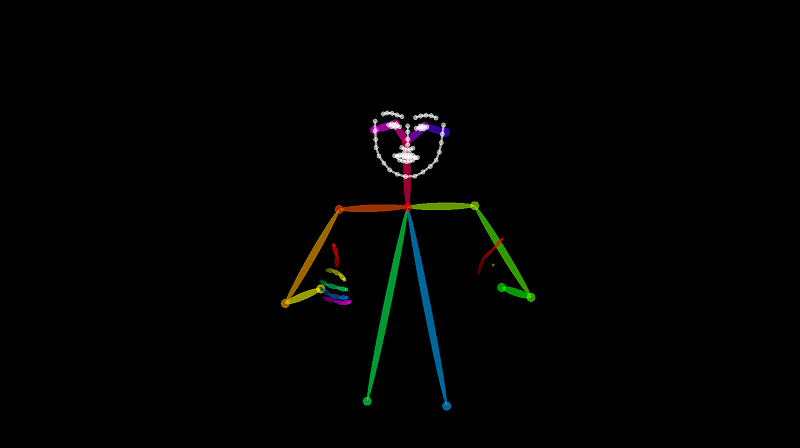}\\
\includegraphics[height=1.69in, width=3.0in]{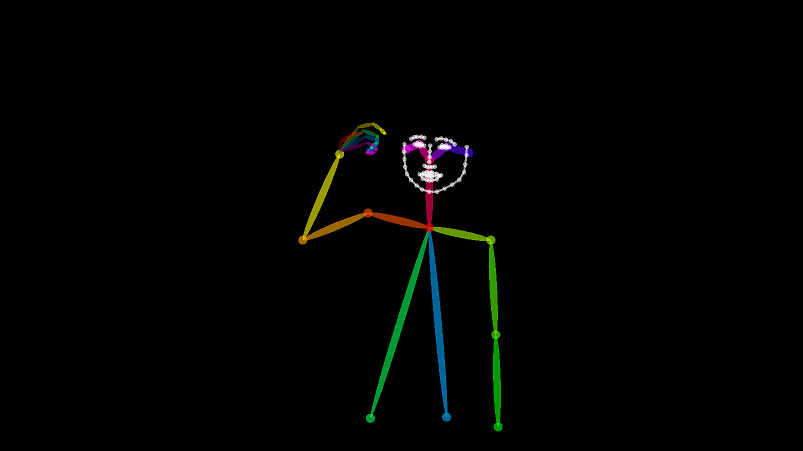}
\includegraphics[height=1.69in, width=3.0in]{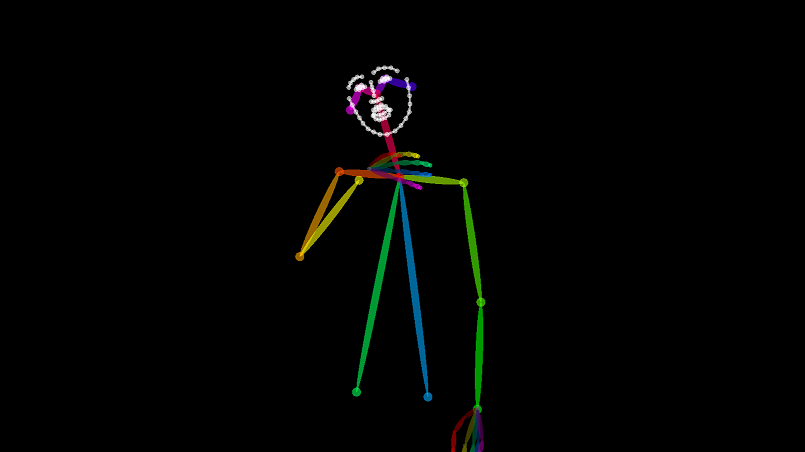}\\
\includegraphics[height=1.69in, width=3.0in]{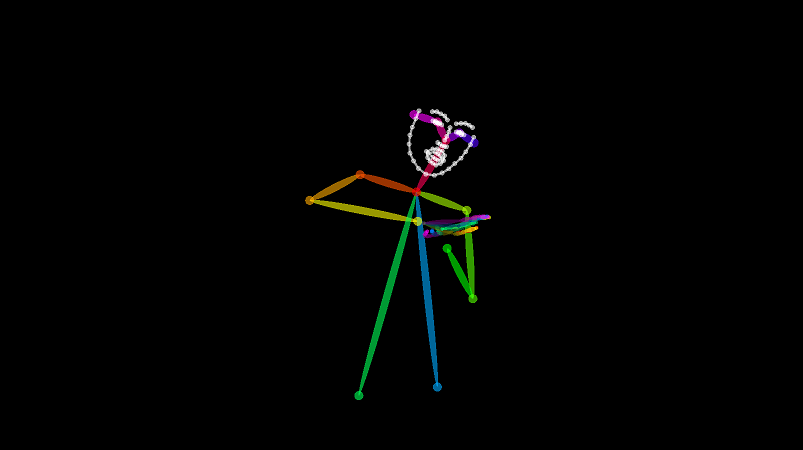}
\includegraphics[height=1.69in, width=3.0in]{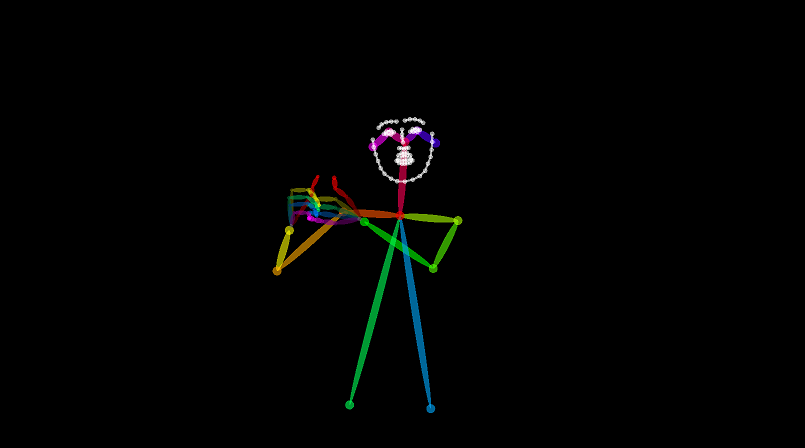}\\
\caption{Keypoints extracted from example frames in Figure~\ref{fig:example_frame2}}
\label{fig:example_frame3}
\end{figure}

The KETI dataset is constructed to understand the Korean sign language of hearing-impaired people in various emergency situations. Indeed, in many social situations, the hearing-impaired people necessarily need help from professional sign language interpreters to communicate with the hearing people even when they have to reveal their very private and sensitive information. Moreover, the hearing-impaired people are more vulnerable in various emergency situations due to the communication barriers due to the absence of the hearing ability.
Therefore, we have carefully examined the cases of relatively general conversations about emergency situations and chosen 105 sentences and 419 words that could be used in various emergency situations.

The KETI sign language dataset consists of 14,672 full high definition (HD) videos, that are recorded at 30 frames per second and from two camera angles; front and side. We have recorded 524 different signs derived from the aforementioned process and performed by fourteen different hearing-impaired signers to reflect the individual differences for the same sign. For each sign, we first record a `guide video' of an `expert' signer to remove the possible ambiguity of signs. After watching the guide video, the fourteen hearing-impaired signers recorded each of the 524 signs.
As a result, each signer records a total of 1,048 videos for the dataset. For the training and validation sets, we have chosen ten signers from fourteen signers and chosen nine sign videos for each sign for the training set. The remaining sign videos are assigned to the validation set. The test set consists of sign videos of four signers who do not appear in any video in the training set or the validation set. Several statistics of the dataset are given in Table \ref{tab:dataset} and an example frame from the dataset is presented in Figure \ref{fig:example_frame}. We also present ten example frames that are extracted from sign videos of ten different signers in Figure \ref{fig:example_frame2}.

In particular, we have annotated each of the 105 signs that correspond to the useful sentences in emergencies mentioned above with five different natural language sentences in Korean.
Moreover, we have annotated all sign videos with the corresponding sequences of {\em glosses}~\cite{Liddell03}, where a gloss is a unique word that corresponds to a unit sign and used to transcribe sign language. For instance, a sign implying `I am burned.' can be annotated with the following sequence of glosses: (`FIRE', `SCAR'). Similarly, a sentence `A house is on fire.' is annotated by (`HOUSE', `FIRE').
Apparently, glosses are more appropriate to annotate a sign because it is possible to be expressed in various natural sentences or words with the same meaning. For this reason, we have annotated all signs with the glosses with the help of Korean sign language experts. 
Table~\ref{tab:annotation} exhibits ten examples from 105 data examples in total.

\begin{table}[!ht]
\centering
{
\begin{tabular}{@{  }clll@{  }} 
\toprule
ID  & Korean Sentence  & English sentence & Sign gloss \\ \midrule
1  & 화상을 입었어요. & I got burned. & FIRE SCAR \\ 
2  & 폭탄이 터졌어요. & The bomb went off. & BOMB \\
3  & 친구가 숨을 쉬지 않아요. & My friend is not breathing. & FRIEND BREATHE CANT\\
4  & 집이 흔들려요. & The house is shaking. & HOUSE SHAKE\\ 
5  & 집에 불이 났어요. & The house is on fire. & HOUSE FIRE\\ 
6  & 가스가 새고 있어요. & Gas is leaking.  & GAS BROKEN FLOW\\
7  & 112에 신고해주세요. & Please call 112. & 112 REPORT PLEASE \\
8  & 도와주세요. & Help me. & HELP PLEASE\\
9  & 너무 아파요. & It hurts so much. & SICK \\
10  & 무릎 인대를 다친 것 같아요. & I hurt my knee ligament. & KNEE LIGAMENT SCAR\\
\bottomrule
\end{tabular}
}
\caption{Ten examples of our sign language annotations. We annotate each sign with five natural language sentences in Korean and a unique sign gloss. We only provide two sentences in the table due to space limitations.}
\label{tab:annotation}
\end{table}

For the communication with hearing-impaired people in the situations, the KETI dataset is used to develop an artificial intelligence-based sign language recognizer or translator. All videos are recorded in a blue screen studio to minimize any undesired influence and learn how to recognize or translate the signs with an insufficient amount of data.

\begin{table}
\centering
\begin{tabular}{@{  }p{4cm}rrr@{  }} 
\toprule
Metric  & Training & Dev & Test\\ \midrule
Number of sign videos  & 9,432 & 1,048 & 4,192  \\ 
Duration [hours]  & 20.05 & 2.24 & 5.70\\
Number of frames  & 2,165,682 & 241,432 & 615,486\\
Number of signers  & 10 & 10 & 4\\ 
Number of camera angles  & \multicolumn{3}{c}{2}\\ 
\bottomrule
\end{tabular}
\caption{Statistics of KETI sign language dataset}
\label{tab:dataset}
\end{table}

\section{Our Approach}
We propose a sign recognition system based on the human keypoints that are estimated by pre-existing libraries such as OpenPose~\cite{CaoSWS17,SimonJMS17,WeiRKS16}. In Figure~\ref{fig:example_frame3}, we provide example results of human keypoint detection by the OpenPose for ten example frames presented in Figure~\ref{fig:example_frame2}. Here we develop our system based on OpenPose, an open source toolkit for real-time multi-person keypoint detection. OpenPose can estimate in total 137 keypoints where 25 keypoints are from body pose, 21 keypoints are from each hand, and 70 keypoints from a face. The primary reason of choosing OpenPose as a feature extractor for sign language recognition is that it is robust to many types of variations.

We use the estimated coordinates of 124 keypoints of a signer to understand the sign language of the signer, where 12 keypoints are from human body, 21 keypoints are from each hand, and 70 keypoints are from face. Note that the number of keypoints from human body is 25 but we select 12 keypoints that correspond to upper body parts. The chosen indices and the names of the parts are as follows: 0 (nose), 1 (neck), 2 (right shoulder), 3 (right elbow), 4 (right wrist), 5 (left shoulder), 6 (left elbow), 7 (left wrist), 15 (right eye), 16 (left eye), 17 (right ear), and 18 (left ear).


\begin{figure}[H]
\centering
\scalebox{0.9}{
\begin{tikzpicture}[
  hid/.style 2 args={
    rectangle split,
    rectangle split horizontal,
    draw=#2,
    rectangle split parts=#1,
    fill=#2!20,
    outer sep=0.5mm,
    inner sep=1.5mm,
    rounded corners}]
  
    \node[inner sep=0.5mm] (i1) at (2.5*1, -6.25) {\includegraphics[height=0.6in, width=0.9in]{frame_86.png}};
    \node[inner sep=0.5mm] (i2) at (2.5*2, -6.25) {\includegraphics[height=0.6in, width=0.9in]{frame_94.png}};
    \node[inner sep=0.5mm] (i3) at (2.5*3, -6.25) {\includegraphics[height=0.6in, width=0.9in]{frame_128.png}};

    \node[inner sep=0.5mm] (i21) at (2.5*1, -3.65) {\includegraphics[height=0.6in, width=0.9in]{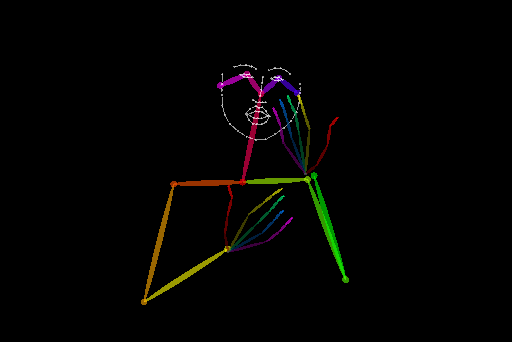}};
    \node[inner sep=0.5mm] (i22) at (2.5*2, -3.65) {\includegraphics[height=0.6in, width=0.9in]{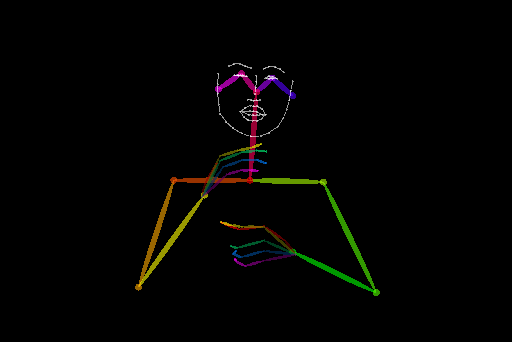}};
    \node[inner sep=0.5mm] (i23) at (2.5*3, -3.65) {\includegraphics[height=0.6in, width=0.9in]{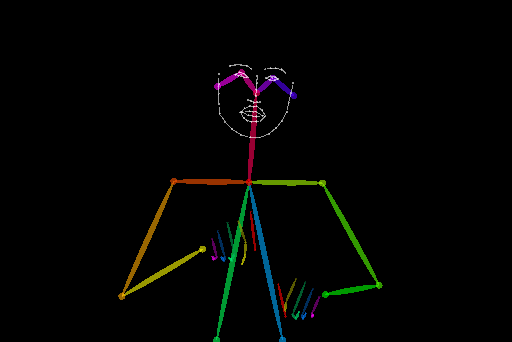}};
  \foreach \t [count=\step from 4] in {I, am,burned,{{$<$eos$>$}}} {
    \node[align=center] (o\step) at (2.5*\step, -1.4) {\t};
  }
  \foreach \step in {1,...,3} {
    \node[hid={1}{red}] (h\step) at (2.5*\step, -0.7) {Enc. GRU};
    \node[hid={1}{red}] (e\step) at (2.5*\step, -1.6) {Enc. GRU};    
    \draw[->] (i\step.north) -> (i2\step.south);
    \draw[->] (i2\step.north) -> (e\step.south);
    \draw[->] (e\step.north) -> (h\step.south);
  }

    \node (i4) at (2.5*4, -5.9) {$<$sos$>$};
    \node (output5) at (2.5*5, -5.9) {I};
    \node (output6) at (2.5*6, -5.9) {am};
    \node (output7) at (2.5*7, -5.9) {burned};
    
  \foreach \step in {4,...,7} {
    \node[hid={1}{yellow}] (s\step) at (2.5*\step, -2.25) {Softmax};
    \node[hid={1}{blue}] (h\step) at (2.5*\step, -3.1) {Dec. GRU};
    \node[hid={1}{blue}] (e\step) at (2.5*\step, -4.0) {Dec. GRU}; 
    \node[hid={1}{orange}] (embedding\step) at (2.5*\step, -5.0) {Embedding};
    \draw[->] (e\step.north) -> (h\step.south);
    \draw[->] (h\step.north) -> (s\step.south);
    \draw[->] (s\step.north) -> (o\step.south);
    \draw[->] (embedding\step.north) -> (e\step.south);
  }  
  \draw[->] (output5.north) -> (embedding5.south);
  \draw[->] (output6.north) -> (embedding6.south);
  \draw[->] (output7.north) -> (embedding7.south);
  
\node[draw=green, fill=green!30, rectangle, minimum width=7cm, rounded corners] at (5, -2.4) {Feature Normalization};
\node[draw=green, fill=green!30, rectangle, minimum width=7cm, rounded corners] at (5, -4.95) {Human Keypoint Estimation};

  \draw[->] (i4.north) -> (embedding4.south);
  \foreach \step in {1,...,2} {
    \pgfmathtruncatemacro{\next}{add(\step,1)}
    \draw[->] (h\step.east) -> (h\next.west);
  }
  \foreach \step in {4,...,6} {
    \pgfmathtruncatemacro{\next}{add(\step,1)}
    \draw[->] (h\step.east) -> (h\next.west);
  }
  \path (h3.east) edge[->,out=-50,in=-240] (h4.west);
  \foreach \step in {4,...,6} {
    \pgfmathtruncatemacro{\next}{add(\step,1)}
    \path (o\step.north) edge[->,out=45,in=225] (output\next.south);
  }
\end{tikzpicture}}
\caption{An overall architecture of our approach that translates a sign language video into a natural language sentence using sequence to sequence model based on GRU cells.}
\label{fig:overall}
\end{figure}

\subsection{Human Keypoint Detection by OpenPose}

First, our recognition system is expected to be robust in different cluttered backgrounds as it only detects the human body. Second, the system based on the human keypoint detection works well regardless of signer since the variance of extracted keypoints are negligible. Moreover, we apply the feature normalization technique to further reduce the variance which is dependent on signer. Third, our system can enjoy the benefits of the improvement on the keypoint detection system which has a great potential in the future because of its versatility. For instance, the human keypoint detection system can be used for recognizing different human behaviors and actions given that the relevant dataset is secured. Lastly, the use of high level features is necessary when the scale of the dataset is not large enough. In the case of sign language dataset, it is more difficult to collect than the other dataset as many professional signers should be utilized for recording sign language videos of high quality. The overall architecture of the proposed system is depicted in Figure~\ref{fig:overall}.

\subsection{Feature Vector Normalization}
There have been many successful attempts to employ various types of normalization methods in order to achieve the stability and speed-up of the training process~\cite{BaKH16,IoffeS15,UlyanovVL16}. One of the main difficulty in sign language translation with the small dataset is the large visual variance as the same sign can look very different depending on the signer. Even if we utilize the feature vector which is obtained by estimating the keypoints of human body, the absolute positions of the keypoints or the scale of the body parts in the frame can be very different.
For this reason, we apply a special normalization method called the {\em object 2D normalization} that suits well in our purpose. 

After extracting high-level human keypoints, we normalize the feature vector using the mean and standard deviation of the vector to reduce the variance of the data. Let us denote a 2D feature vector by $V = (v_1, v_2, \ldots, v_n) \in \mathbb{N}^{n\times2}$ that consists of $n$ elements where each element~$v_i \in \mathbb{N}^2,\;1 \le i \le n$ stands for a single keypoint of human part. Each element $v_i = (v_i^x, v_i^y)$ consists of two integers~$v_i^x$ and $v_i^y$ that imply the $x$- and the $y$-coordinates of the keypoint~$v_i$ in the video frame, respectively. From the given feature vector $V$, we can extract the two feature vectors as follows:
\[
V_x = (v_1^x, v_2^x, \ldots, v_n^x)\textrm{ and }
V_y = (v_1^y, v_2^y, \ldots, v_n^y).
\]
Simply speaking, we collect the $x$ and $y$-coordinates of keypoints separately while keeping the order. Then, we normalize the $x$-coordinate vector~$V_x$ as follows:
\[
V_x^* = \frac{V_x - \bar{V_x}}{\sigma(V_x)},
\]
where $\bar{V_x}$ is the mean of $V_x$ and $\sigma(V_x)$ is the standard deviation of $V_x$. 
Note that $V_y^*$ is calculated analogously.
Finally, it remains to concatenate the two normalized vectors to form the 
final feature vector $V^* = [V_x^*; V_y^*] \in \mathbb{N}^{2n}$ which will 
be used as the input vector of our neural network.

It should be noted that we assume that the keypoints of lower body parts are not necessary for sign language recognition. Therefore, we only use 124 keypoints from the 137 keypoints detected by OpenPose since six keypoints of human pose correspond to lower body parts such as both feet, knees and pelvises as you can see in Figure 2. We randomly sample 10 to 50 keyframes from each sign video. Hence, the dimension of input feature vector is $248 \times |V|$, where $|V| \in \{ 10, 20, 30, 40, 50\}$.

\subsection{Frame Skip Sampling for Data Augmentation}

The main difficulty of training neural networks with small datasets is that
the trained models do not generalize well with data from the validation and 
the test sets. As the size of dataset is even smaller than the usual cases 
in our problem, we utilize the {\em random frame skip sampling} that is commonly used to 
process video data such as video classification~\cite{KarpathyTSLSF14} for augmenting 
training data. The effectiveness if data augmentation has been proved in 
many tasks including image classification~\cite{PerezW17}. Here, we randomly extract multiple representative features of a video. 

Given a sign video~$S = (f_1, f_2, \ldots, f_l )$ that contains $l$ frames from $f_1$ to $f_l$, we randomly select a fixed number of frames, say $n$. Then, we first compute the average length of gaps between frames as follows:
\[
z = \left\lfloor\dfrac{l}{n-1}\right\rfloor.
\]

We first extract a sequence of frames with indices from the following sequence $Y =(y, y +z, y+ 2z \ldots, y + (n-1)z ) \in \mathbb{N}^n$, where $y = \lfloor\frac{l - z(n-1)}{2}\rfloor$ and call it a {\em baseline sequence}. Then, we generate a random integer sequence $R = (r_1,r_2, \ldots, r_n) \in [1,z]^n$ and compute the sum 
of the random sequence and the baseline sequence. Note that the value of the last index is clipped to the value in the range of $[1,l]$. We start from the baseline sequence instead of choosing any random sequence of length $l$ to avoid generating random sequences of frames that are possibly not containing `key' moments of signs.

\subsection{Attention-based Encoder-Decoder Network}

The encoder-decoder framework based on RNN architectures such as LSTMs or GRUs is gaining its popularity for neural machine translation~\cite{BahdanauCB14,LuongPM15,SutskeverVL14,VaswaniSPUJGKP17} as it successfully replaces the statistical machine translation methods.

Given an input sentence ${\bf x} = (x_1, x_2, \ldots, x_{T_x})$, an 
encoder RNN plays its role as follows:
\[
h_t = {\rm RNN}(x_t, h_{t-1})
\]
where $h_t \in \mathbb{R}^n$ is a hidden state at time $t$. After processing the whole input sentence, the encoder generates a fixed-size context vector that represents the  sequence as follows:
\[
c = q(h_1, h_2, \ldots, h_{T_x}),
\]
 
For instance, the RNN is an LSTM cell and $q$ simply returns the last hidden state $h_{T_x}$ in one of the original sequence to sequence paper by Sutskever et al.~\cite{SutskeverVL14}. 

Now suppose that ${\bf y} = (y_1, y_2, \ldots, y_{T_y})$ is an output sentence that corresponds to the input sentence ${\bf x}$ in training set. Then, the decoder RNN is trained to predict the next word conditioned on all the previously predicted words and the context vector from the encoder RNN. In other words, the decoder computes a probability of the translation {\bf y} by decomposing the joint probability into the ordered conditional probabilities as follows:
\[
p({\bf y}) = \prod_{i=1}^{T_y} p(y_i | \{ y_1, y_2, \ldots, y_{i-1}\}, c).
\]

Now our RNN decoder computes each conditional probability as follows:
\[
p(y_i | y_1, y_2, \ldots, y_{i-1}, c) = {\rm softmax}(g(s_{i})),
\]
where $s_i$ is the hidden state of decoder RNN at time $i$ and $g$ is a linear transformation 
that outputs a vocabulary-sized vector.
Note that the hidden state~$s_i$ is computed by
\[
s_i = {\rm RNN}(y_{i-1},s_{i-1}, c),
\]
where $y_{i-1}$ is the previously predicted word, $s_{i-1}$ is the last hidden state of decoder RNN, and $c$ is the context vector computed from encoder RNN.\\

\noindent{\bf Bahdanau attention.} Bahdanau et al.~\cite{BahdanauCB14} conjectured that the fixed-length context vector $c$ is a bottleneck in improving the performance of the translation model and proposed to compute the context vector by automatically searching for relevant parts from the hidden states of encoder. Indeed, this `attention' mechanism has proven really useful in various tasks including but not limited to machine translation. They proposed a new model that defines each conditional probability at time $i$ depending on a dynamically computed context vector $c_i$ as follows:
\[
p(y_i | y_1, y_2, \ldots, y_{i-1}, {\bf x}) = {\rm softmax}(g(s_i)),
\]
where $s_i$ is the hidden state of the decoder RNN at time $i$ which is computed by
\[
s_i = {\rm RNN}(y_{i-1}, s_{i-1}, c_i).
\]

The context vector $c_i$ is computed as a weighted sum of the hidden states from encoder:
\[
c_i = \sum_{j=1}^{T_x} \alpha_{ij} h_j,
\]
where
\[
\alpha_{ij} = \frac{\exp ({\rm score}(s_{i-1}, h_j))}{\sum_{k=1}^{T_x} \exp({\rm score}(s_{i-1}, h_k))}.
\]

Here the function `score' is called an {\em alignment function} that computes how well the two hidden states from the encoder and the decoder, respectively, match. For example, ${\rm score}(s_i, h_j)$, where $s_i$ is the hidden state of the encoder at time $i$ and $h_j$ is the hidden state of the decoder at time $j$ implies the probability of aligning the part of the input sentence around position $i$ and the part of the output sentence around position $j$.
\\

\noindent{\bf Luong attention.} Later, Luong et al.~\cite{LuongPM15} examined a novel attention mechanism which is very similar to the attention mechanism by Bahdananu et al. but different in some details. First, only the hidden states of the top RNN layers in both the encoder and decoder are used instead of using the concatenation of the forward and backward hidden states of the bi-directional encoder and the hidden states of the uni-directional non-stacking decoder. Second, the computation path is simplified by computing the attention matrix after computing the hidden state of the decoder at current time step. They also proposed the following three scoring functions to compute the degree of alignment between the hidden states as follows:
\[
{\rm score}(h_t, h_s) = 
\begin{cases}
h_t^\intercal h_s, & \textrm{(Dot)}\\
h_t^\intercal W h_s, & \textrm{(General)} \\ 
V^\intercal \tanh(W [h_t ;h_s]), & \textrm{(Concat.)}
\end{cases}
\]
where $V$ and $W$ are learned weights. Note that the third one based on the concatenation is originally proposed by Bahdanau et al.~\cite{BahdanauCB14}.
\\

\noindent{\bf Multi-head attention (Transformer).} While the previous encoder-decoder architectures are based on RNN cells, Vaswani et al.~\cite{VaswaniSPUJGKP17} proposed 
a completely new network architecture which is based solely on attention mechanisms 
without any recurrence and convolutions. The most important characteristic of the Transformer is the {\em multi-head attention} which is used in three different ways as follows:
\begin{enumerate}
    \item Encoder-decoder attention: each position in the decoder can attend over all positions in the input sequence.
    \item Encoder self-attention: each position in the encoder can attend over all positions in the previous layer of the encoder.
    \item Decoder self-attention: each position in the decoder can attend over all positions in the decoder up to and that position.
\end{enumerate}

Moreover, as the Transformer uses neither recurrence nor convolution, the model requires some information about the order of the sequence. To cope with this problem, the Transformer uses {\em positional encoding} which contains the information about the relative or absolute position of the words in the sequence using sine and cosine functions.

\begin{table}
\centering
\begin{tabular}{@{ }rcccccccc@{ }}
\toprule
& \multicolumn{4}{c}{Validation Set} & \multicolumn{4}{c}{Test Set} \\
\cmidrule(lr){2-5} \cmidrule(lr){6-9}
Attention type    &   ROUGE-L & METEOR  & BLEU  & CIDEr &  ROUGE-L & METEOR  & BLEU & CIDEr\\
\midrule
Vanilla seq2seq~\cite{SutskeverVL14} & 90.03 & 62.66 & 87.79 & 3.838 & 62.93 & 38.03 & 50.80 & 2.129 \\
Bahdanau et al.~\cite{BahdanauCB14}  & 94.72 & 67.44 & 94.03 & 4.264 & 71.45 & 44.06 & 63.38 & 2.616 \\
Luong et al.~\cite{LuongPM15}        & \textbf{96.63} & \textbf{72.24} & \textbf{95.86} & \textbf{4.322} & \textbf{73.61} & 46.52 & 65.26 & 2.794 \\
Transformer~\cite{VaswaniSPUJGKP17}  & 94.14 & 69.27 & 92.90 & 4.227 & 73.18 & \textbf{47.03} & \textbf{66.58} & \textbf{2.857}\\ 
\bottomrule
\end{tabular}
\caption{Performance comparison of sign language translation on different types of attention mechanisms.}
\label{tab:attention}
\end{table}

\section{Experimental Results}

We implemented our networks using PyTorch~\cite{PaszkeGCCYDLDAL17}, which is an open source machine learning library for Python.
The Adam optimizer~\cite{KingmaB14} was used to train the network weights and biases for 50 epochs with an initial learning rate $0.001$. During the training, we changed the learning rate every 20 epochs by the exponential decay scheduler with discount factor~$0.5$. We also used the dropout regularization with a probability of 0.8 and the gradient clipping with a threshold 5. Note that the dropout regularization is necessarily high as the size and the variation of the dataset is small compared to other datasets specialized for deep learning training. For the sequence-to-sequence models including the vanilla seq2seq model and two attention-based models, the dimension of hidden states is 256. For the Transformer model, we use the dimension for input and output ($d_{\rm model}$ in~\cite{VaswaniSPUJGKP17}) of 256. The other hyper-parameters used for the Transformer are the same as in the original model including the scheduled Adam optimizer in their own setting. Moreover, the batch size is 128, the augmentation factor is 100, the number of chosen frames is 50, and the object 2D normalization is used unless otherwise specified.

As our dataset is annotated in Korean which is an agglutinative language, the morphological analysis on the annotated sentences should be performed because the size of dictionary can be arbitrarily large if we split sentences into words simply by white-spaces in such languages. For this reason, we used the Kkma part-of-speech (POS) tagger in the KoNLPy package which is a Python package developed for natural language processing of the Korean language to tokenize the sentences into the POS level~\cite{ParkC14}.

In order to evaluate the performance of our translation model, we basically calculate `accuracy' which means the ratio of correctly translated words and sentences. It should be mentioned that the accuracy is calculated only for comparing sentence-level annotation and gloss-level annotation in Table~\ref{tab:sentence_gloss} since it is not suitable to compare two cases by accuracy since we have five ground truth sentences for each sing video. In this experiment we train our model with a single ground truth sentence or a single sequence of glosses. Besides, we also utilized three types of metrics that are commonly used for measuring the performance of machine translation models such as BLEU~\cite{PapineniRWZ02}, ROUGE-L~\cite{Lin08}, METEOR~\cite{BanerjeeL05}, and CIDEr~\cite{VedantamZP15} scores.
\\

\noindent{\bf Sentence-level vs Gloss-level training.} As in~\cite{CamgozHKNB18}, we conduct an experiment to compare the translation performance depending on the type of annotations. Because each sign corresponds to a unique sequence of glosses while it corresponds to multiple natural language sentences, it is easily predictable that the gloss-level translation shows better performance. Indeed, we can confirm the anticipation from the summary of results provided in Table~\ref{tab:sentence_gloss}. 

This also leads us to the future work for translating sequences of glosses into natural language sentences. We expect that the sign language translation can be a more feasible task by separating the task of annotating sign videos with natural language sentences by two sub-tasks where we annotate sign videos with glosses and annotate each sequence of glosses with natural language sentences.\\

\begin{table}[!htb]
\centering
\begin{tabular}{@{ } rcccccccc @{ }}
\toprule
& \multicolumn{4}{c}{Validation Set} & \multicolumn{4}{c}{Test Set} \\
\cmidrule(lr){2-5} \cmidrule(lr){6-9}
Annotation   &  Accuracy & ROUGE-L & METEOR  & BLEU  & Accuracy & ROUGE-L & METEOR  & BLEU \\
\midrule
Sentence-level &  82.07 & 94.42 & 67.35 & 90.57 & 45.56 & \textbf{66.10} & \textbf{41.09} & \textbf{57.37} \\
Gloss-level    &  \textbf{93.28} & \textbf{96.03} & \textbf{71.04} & \textbf{93.85} & \textbf{55.28} & 63.53 & 38.10 & 52.63 \\
\bottomrule
\end{tabular}
\caption{Comparison of sign language translation performance on different types of annotations.}
\label{tab:sentence_gloss}
\end{table}

\noindent{\bf Comparison with CNN-based approaches.}
In Table~\ref{tab:CNN}, we compare our approach to the classical methods 
based on CNN features extracted from well-known architectures such as 
ResNet~\cite{HeZRS16} and VGGNet~\cite{SimonyanZ14a}.
Since the size of sign video frames ($1,920 \times 1,080$) is different to the size of input of CNN models (224 $\times$ 224), we first crop the central area of frames in $1,080 \times 1,080$ and resize the frames to $224 \times 224$.

The experimental results show that ResNet-152 exhibits the best translation performance on the validation set and the VGGNet-16 demonstrates the best performance on the test set. In general, the performance difference on the validation set is not large but it is apparent that the VGGNet models are much better in generalizing to the test set compared to the ResNet models. 

Expectably, the translation models using the CNN extracted features show significantly worse translation performance than the models using the human keypoint features. It is well-known that CNN-based architectures such as ResNet and VGGNet have a huge number of trainable parameters (e.g., VGGNet-19 and ResNet-152 have over 143M and 60M parameters, respectively.) so that they easily fall into the overfitting problem due to the lack of a sufficient number of training examples. Moreover, the CNN-based models have a weakness for recognizing signs of previously unseen signers as they are weaker than our model in dealing with subtle variances of images.

It is still interesting to know whether the combination of any CNN-based features and human keypoint features works better than when we solely rely on the human keypoint features. As the size of sign language dataset grows, we expect that the CNN-based models improve their performances and generalize much better.
\\

\begin{table}[!htb]
\centering
\begin{tabular}{@{ }rcccccccc@{ }}
\toprule
& \multicolumn{4}{c}{Validation Set} & \multicolumn{4}{c}{Test Set} \\
\cmidrule(lr){2-5} \cmidrule(lr){6-9}
Feature type    &   ROUGE-L & METEOR  & BLEU  & CIDEr &  ROUGE-L & METEOR  & BLEU & CIDEr\\
\midrule
VGGNet-16~\cite{SimonyanZ14a}  & 66.85 & 41.92 & 56.75 & 2.369 & \textbf{44.75} & \textbf{25.79} & \textbf{27.88} & \textbf{1.016} \\ 
VGGNet-19~\cite{SimonyanZ14a}  & 61.77 & 38.72 & 50.95 & 2.060 & 42.75 & 24.81 & 24.27 & 0.839 \\
ResNet-50~\cite{HeZRS16}       & 62.26 & 38.79 & 51.99 & 2.124 & 38.76 & 21.85 & 19.45 & 0.664 \\
ResNet-101~\cite{HeZRS16}      & 66.28 & 41.81 & 56.26 & 2.368 & 40.10 & 22.68 & 21.86 & 0.772 \\
ResNet-152~\cite{HeZRS16}      & \textbf{74.10} & \textbf{48.03} & \textbf{66.73} & \textbf{2.841} & 38.44 & 22.71 & 20.78 & 0.753 \\
\midrule
OpenPose~\cite{CaoSWS17,SimonJMS17,WeiRKS16} & 96.92 & 72.14 & 96.11 & 4.380 & 73.95 & 46.66 & 64.79 & 2.832 \\
\bottomrule
\end{tabular}
\caption{Performance comparison with translation models based on CNN-based feature extraction techniques. Note that the augmentation factor in this experiment is all set to 50.}
\label{tab:CNN}
\end{table}

\noindent{\bf Effect of feature normalization methods.} 
In order to evaluate the effect of the feature normalization method on the keypoints estimated by OpenPose, we compare the following five cases: 1) no normalization, 2) feature normalization, 3) object normalization, 4) 2-dimensional (2D) normalization, and 5) object 2D normalization. In the first case, we do not perform any normalization step on the keypoint feature generated by concatenating the coordinate values of all keypoints. In the feature normalization, we create a keypoint feature as in 1) and normalize the feature with the mean and standard deviation of the whole feature. In the object normalization, we normalize the keypoint features obtained from two hands, body, and face, respectively, and concatenate them to generate a feature that represents the frame. We also consider the case of 2D normalization in which we normalize the $x$- and $y$-coordinates separately.
Lastly, the object 2D normalization is the normalization method that we propose in the paper.

\begin{table}[htb!]
\centering
{\small 
\begin{tabular}{@{  }p{4cm}cccc@{  }} 
\toprule
Method & ROUGE-L & METEOR & BLEU & CIDEr \\ \midrule
Feature Normalization   & 66.28 & 40.94 & 56.91 & 2.401 \\
2D Normalization        & 72.05 & 44.98 & 62.69 & 2.706 \\
Object Normalization    & 64.16 & 38.84 & 53.83 & 2.235 \\
Object 2D Normalization & \textbf{73.61} & \textbf{46.52} & \textbf{65.26} & \textbf{2.794} \\
\bottomrule
\end{tabular}
}
\caption{Effect of different feature normalization methods on the translation performance. The results are obtained on the test set.}
\label{tab:normalization}
\end{table}

Table~\ref{tab:normalization} summarizes the result of our experiments.
The table does not contain the results of the case without any normalization as it turns out that the proposed object 2D normalization method is superior to the other normalization methods we considered. Especially, when we train our neural network with the keypoint feature vector which is obtained by simply concatenating the $x$ and $y$ coordinates of keypoints without any normalization, the validation loss never decreases. While any kind of normalization seems working positively, it is quite interesting to see that there is an additional boost in translation performance when the object-wise normalization and the 2D normalization are used together.
\\

\begin{table}[htb!]
\centering
\begin{tabular}{@{ }rcccccccc@{ }}
\toprule
& \multicolumn{4}{c}{Validation Set} & \multicolumn{4}{c}{Test Set} \\
\cmidrule(lr){2-5} \cmidrule(lr){6-9}
Augmentation factor    &   ROUGE-L & METEOR  & BLEU  & CIDEr &  ROUGE-L & METEOR  & BLEU & CIDEr\\
\midrule
100 & 96.63 & \textbf{72.24} & 95.86 & 4.322 & 73.61 & 46.52 & \textbf{65.26} & 2.794 \\
50  & \textbf{96.92} & 72.14 & \textbf{96.11} & \textbf{4.380} & \textbf{73.95} & \textbf{46.66} & 64.79 & \textbf{2.832} \\
10  & 95.69 & 70.14 & 94.46 & 4.227 & 71.40 & 45.10 & 62.95 & 2.662 \\
\bottomrule
\end{tabular}
\caption{Effects of data augmentation by random frame sampling on sign language translation performance.}
\label{tab:augmentation}
\end{table}

\begin{table}
\centering
\begin{tabular}{@{ }rcccccccc@{ }}
\toprule
& \multicolumn{4}{c}{Validation Set} & \multicolumn{4}{c}{Test Set} \\
\cmidrule(lr){2-5} \cmidrule(lr){6-9}
Number of frames    &   ROUGE-L & METEOR  & BLEU  & CIDEr &  ROUGE-L & METEOR  & BLEU & CIDEr\\
\midrule
50 & \textbf{96.63} & 72.24 & 95.86 & 4.322 & \textbf{73.61} & \textbf{46.52} & \textbf{65.26} & \textbf{2.794} \\
40 & 96.52 & \textbf{72.36} & \textbf{95.96} & \textbf{4.327} & 72.71 & 46.18 & 64.35 & 2.757 \\
30 & 95.88 & 70.60 & 94.87 & 4.281 & 73.35 & 46.46 & 64.48 & 2.778 \\
20 & 94.38 & 68.40 & 92.98 & 4.181 & 72.37 & 45.37 & 62.19 & 2.693 \\
10 & 83.26 & 55.81 & 78.43 & 3.427 & 65.89 & 40.65 & 54.01 & 2.308 \\
\bottomrule
\end{tabular}
\caption{Effects of the number of sampled frames on sign language translation performance.}
\label{tab:numframes}
\end{table}

\begin{table}[htb!]
\centering
\begin{tabular}{@{ }rcccccccc@{ }}
\toprule
& \multicolumn{4}{c}{Validation Set} & \multicolumn{4}{c}{Test Set} \\
\cmidrule(lr){2-5} \cmidrule(lr){6-9}
Batch size    &   ROUGE-L & METEOR  & BLEU  & CIDEr &  ROUGE-L & METEOR  & BLEU & CIDEr\\
\midrule
128 & 96.63 & 72.24 & 95.86 & 4.322 & \textbf{73.61} & \textbf{46.52} & \textbf{65.26} & \textbf{2.794} \\
64  & \textbf{96.94} & \textbf{73.20} & \textbf{96.35} & \textbf{4.332} & 72.93 & 46.06 & 63.91 & 2.725 \\
32  & 95.65 & 70.68 & 94.57 & 4.231 & 71.94 & 45.30 & 62.71 & 2.673 \\
16  & 93.74 & 67.58 & 92.74 & 4.118 & 70.63 & 43.86 & 61.94 & 2.571 \\
\bottomrule
\end{tabular}
\caption{Effects of the batch size on sign language translation performance.}
\label{tab:batch_size}
\end{table}

\noindent{\bf Effect of attention mechanisms.}
Here we compare four types of encoder-decoder architectures that are specialized in various machine translation tasks. Table~\ref{tab:attention} demonstrates the clear contrast between the attention-based model by Luong et al.~\cite{LuongPM15} and the Transformer~\cite{VaswaniSPUJGKP17}. While the model of Luong et al. shows better performance than the Transformer on the validation set that contains more similar data to the training set, the Transformer generalizes much better to the test set which consists of sign videos of an independent signer.\\

\noindent{\bf Effect of augmentation factor.} 
We examine the effect of data augmentation by random frame skip sampling and summarize the experimental results in Table~\ref{tab:augmentation}. We call the number of training samples randomly sampled from a single sign video the {\em augmentation factor}. Since the number of sign videos in the training set is 9,432, the total number of training samples after the data augmentation is 943,200 when the augmentation factor is 100.

It should be noted that we do not include the result when we do not augment data by random frame sampling because the validation loss does not decrease at all due to severe overfitting. The result shows that the optimal augmentation factor is 50 for the validation and test set. This implies that the larger augmentation 
factor does not always lead to improvement in performance and even in generalization capability.\\

\noindent{\bf Effect of the number of sampled frames.} 
It is useful to know the optimal number of frames if we plan to develop a real-time sign language translation system because we can reduce the computational cost of the inference engine by efficiently skipping unnecessary frames.
Table~\ref{tab:numframes} shows how the number of sampled frames affects the translation performance. As the sequence-to-sequence model works for any variable-length input sequences, we do not necessarily fix the number of sampled frames. However, it is useful to know the optimal number of frames as the translation performance of the sequence-to-sequence models tends to decline with longer input sequences due to the vanishing gradient problem~\cite{PascanuMB13}. 
Our experimental result shows that the optimal number of frames for the best translation performance is 40 for the validation set and 50 for the test set.
\\

\noindent{\bf Effect of batch size.} 
Recently, it is increasingly accepted that training with small batch often generalizes better to the test set than training with large batch~\cite{HofferHS17,SmithKL17}. However, our experimental results provided in
Table~\ref{tab:batch_size} shows the opposite phenomenon. We suspect that this is due to the scale of the original dataset because large batch is known to be useful to prevent overfitting to some extent.
\\

\begin{figure}[htb!]
\centering
\includegraphics[height=1.69in, width=3.0in]{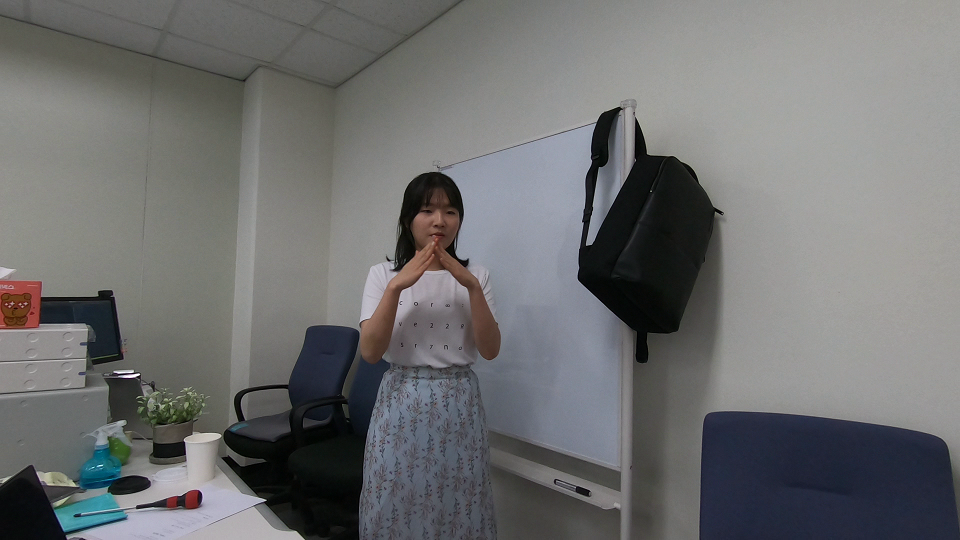}
\includegraphics[height=1.69in, width=3.0in]{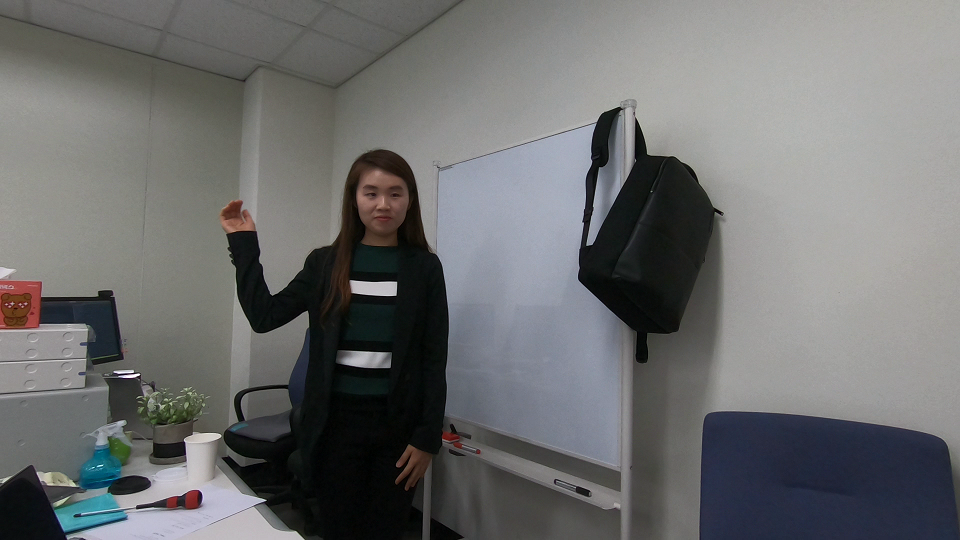}\\
\includegraphics[height=1.69in, width=3.0in]{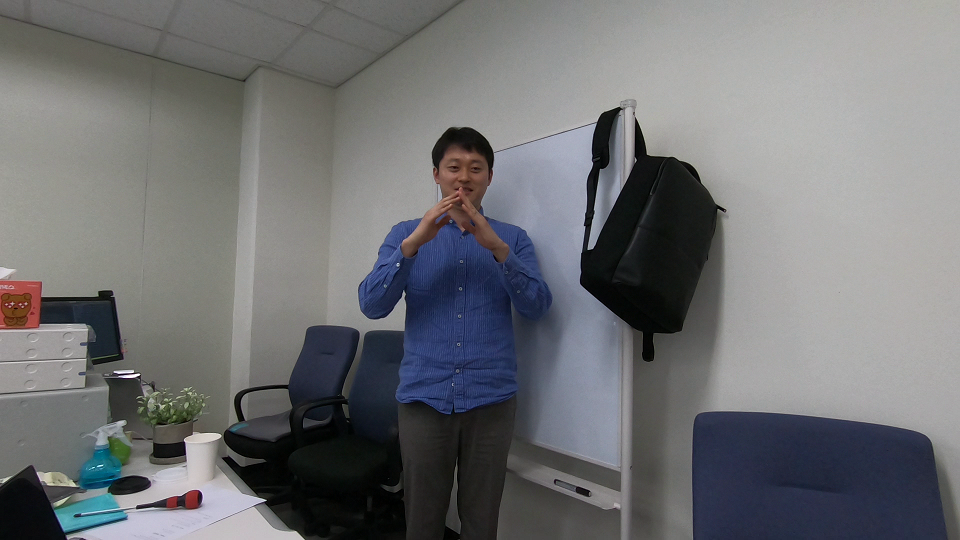}
\includegraphics[height=1.69in, width=3.0in]{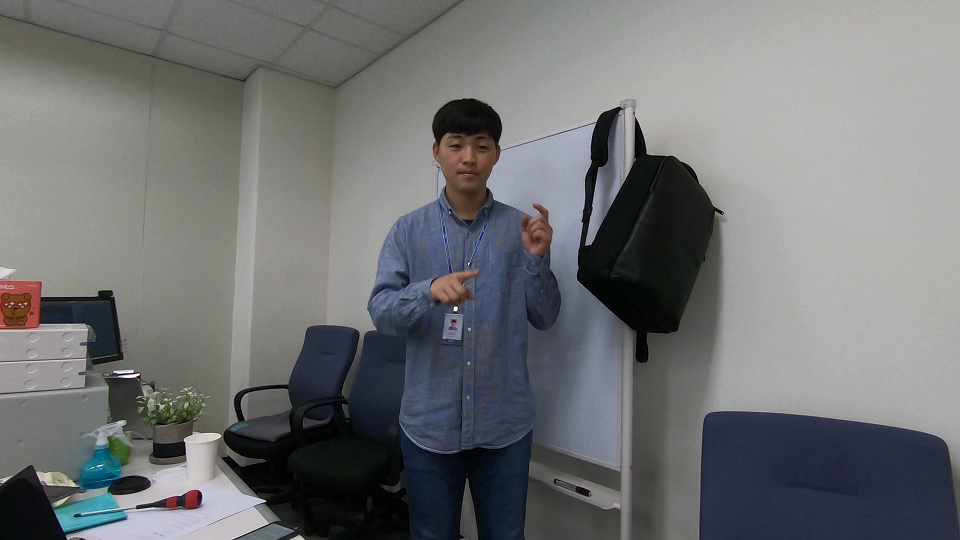}
\caption{Example frames of additional test set videos recorded with cluttered background and non-expert signers}
\label{fig:example_frame4}
\end{figure}

\noindent{\bf Generalizations to real world data.} As we can see in Figure~\ref{fig:example_frame2}, the sign videos in our dataset are recorded in a clear background. This leads us to investigate the performance of our model when the background area of sign videos is cluttered. Moreover, we also check how well our system generalizes to the real world situations by testing the system against more realistic sign video examples that are collected in a much less constrained setting. We collected 30 additional sign videos of five signs from six novice signers in relatively cluttered background to test the generalization ability of our system. Note that we selected relatively easier five signs among the 105 signs since it is very difficult to follow complicated signs for the novice singers who never learned signs before.  Figure~\ref{fig:example_frame4} contains example frames from the additional sign videos. As a result, our system achieves 89.06 (ROUGE-L), 60.64 (METEOR), 77.08 (BLEU) and 2.860 (CIDEr) in the four metrics. The experimental result shows that our system generalizes well to the real world conditions such as cluttered background area and clumsy signs from non-experts.

\subsection{Ablation study}

We also study the effect of the use of keypoint information from two hands, body, and face. The experimental results summarized in Table~\ref{tab:ablation} imply that the keypoint information from both hands is the most important among all the keypoint information from hands, face, and body. 

\begin{table}[htb!]
\centering
{\small
\begin{tabular}{@{  }p{3cm}cccc@{  }} 
\toprule
Method & ROUGE-L & METEOR & BLEU & CIDEr \\ \midrule
Body             & 68.82 & 42.68 & 59.96 & 2.554 \\
Hand             & 72.03 & 45.19 & 62.90 & 2.681 \\
Body, Face       & 60.08 & 36.66 & 48.90 & 2.051 \\
Hand, Face       & 69.43 & 43.26 & 59.92 & 2.538 \\
Hand, Body       & \textbf{74.02} & \textbf{46.84} & \textbf{65.83} & \textbf{2.831} \\
Hand, Body, Face & 73.61 & 46.52 & 65.26 & 2.794 \\
\bottomrule
\end{tabular}
}
\caption{Ablation study on the contributions of keypoints from body, face, and hands. The results are obtained on the test set.}
\label{tab:ablation}
\end{table}

Interestingly, the experimental result tells us that the keypoint information from face does not help to improve the performance in general. The performance even drops when we add face keypoints in all cases. We suspect that the reason is partly due to the imbalanced number of keypoints from different parts. Recall that the number of keypoints from face is 70 and this is much larger than the number of the other keypoints.

While the keypoints from both hands are definitely the most important features to understand signs, it is worth noting that the 12 keypoints from body are boosting up the performance. Actually, we lose the information about relative positions of parts from each other as we normalize the coordinates of each part separately. For instance, there is no way to infer the relative positions of two hands with the normalize feature vectors from both hands. However, it is possible to know the relative position from the keypoints of body as there also exist keypoints corresponding to the hands.

\section{Conclusions}
In this work, we have introduced a new sign language dataset which is manually annotated in Korean spoken language sentences and proposed a neural sign language translation model based on the sequence-to-sequence translation models. It is well-known that the lack of large sign language dataset significantly hinders the full utilization of neural network based algorithms for the task of sign language translation that are already proven very useful in many tasks. Moreover, it is really challenging to collect a sufficient amount of sign language data as we need helps from sign language experts. 
For this reason, we claim that it is inevitable to extract high-level features from sign language video with a sufficiently lower dimension. We are able to successfully train a novel sign language translation system based on the human keypoints that are estimated by a famous open source project called OpenPose~\cite{SimonJMS17,WeiRKS16,CaoSWS17} developed by Hidalgo et al.

In the future, we aim at improving our sign language translation system by exploiting various data augmentation techniques using the spatial properties of videos. We also expect that the performance of the proposed system can be improved if the performance of the human keypoint detection is improved. For instance, there have been various approaches in human keypoint detection such as Mask R-CNN~\cite{HeGDG17} and AlphaPose~\cite{FangXTL17} that exhibit even better performances than OpenPose in terms of accuracy. It is also possible to apply landmark detection methods~\cite{DuanSBDBDDOR18,DuanBSBTCAGOR19,WuJ19} for better performance of keypoint detection from human body, face and both hands. We plan to implement different keypoint detection methods for sign language translation and compare the performances of the methods. It is also important to expand the KETI sign language dataset to sufficiently larger scale by recording videos of more signers in different environments.

\section*{Acknowledgements}

We thank the anonymous referees for a careful reading of an earlier version of the paper and for many valuable suggestions that have improved the presentation. We also sincerely thank our colleagues from Korea Nazarene University who provided insight and expertise in Korean sign language that greatly assisted our research.

This work was supported by the IT R\&D program of MSIT/IITP [2017-0-00255, Autonomous digital companion framework and application].

\clearpage

{\small
\bibliographystyle{ieee}
\bibliography{egbib}

\begin{thebibliography}{10}\itemsep=-1pt

\bibitem{BaKH16}
L.~J. Ba, R.~Kiros, and G.~E. Hinton.
\newblock Layer normalization.
\newblock {\em CoRR}, abs/1607.06450, 2016.

\bibitem{BahdanauCB14}
D.~Bahdanau, K.~Cho, and Y.~Bengio.
\newblock Neural machine translation by jointly learning to align and
  translate.
\newblock In {\em Proceedings of the 3rd International Conference on Learning
  Representations (ICLR 2015)}, 2015.

\bibitem{BanerjeeL05}
S.~Banerjee and A.~Lavie.
\newblock Meteor: An automatic metric for mt evaluation with improved
  correlation with human judgments.
\newblock In {\em Proceedings of the ACL Workshop on Intrinsic and Extrinsic
  Evaluation Measures for Machine Translation and/or Summarization}, pages
  65--72, 2005.

\bibitem{BuehlerZE09}
P.~Buehler, A.~Zisserman, and M.~Everingham.
\newblock Learning sign language by watching {TV} (using weakly aligned
  subtitles).
\newblock In {\em Proceedings of 2009 IEEE Conference on Computer Vision and
  Pattern Recognition (CVPR 2009)}, pages 2961--2968, 2009.

\bibitem{CaoSWS17}
Z.~Cao, T.~Simon, S.~Wei, and Y.~Sheikh.
\newblock Realtime multi-person 2d pose estimation using part affinity fields.
\newblock In {\em Proceedings of 2017 {IEEE} Conference on Computer Vision and
  Pattern Recognition (CVPR 2017)}, pages 1302--1310, 2017.

\bibitem{ChovGBBSB14}
K.~Cho, B.~van Merrienboer, C.~Gulcehre, D.~Bahdanau, F.~Bougares, H.~Schwenk,
  and Y.~Bengio.
\newblock Learning phrase representations using rnn encoder--decoder for
  statistical machine translation.
\newblock In {\em Proceedings of the 2014 Conference on Empirical Methods in
  Natural Language Processing (EMNLP 2014)}, pages 1724--1734, 2014.

\bibitem{CamgozHKNB18}
N.~Cihan~Camgoz, S.~Hadfield, O.~Koller, H.~Ney, and R.~Bowden.
\newblock Neural sign language translation.
\newblock In {\em Proceedings of 2018 IEEE Conference on Computer Vision and
  Pattern Recognition (CVPR 2018)}, pages 7784--7793, 2018.

\bibitem{CooperB09}
H.~Cooper and R.~Bowden.
\newblock Learning signs from subtitles: A weakly supervised approach to sign
  language recognition.
\newblock In {\em Proceedings of 2009 IEEE Conference on Computer Vision and
  Pattern Recognition (CVPR 2009)}, pages 2568--2574, 2009.

\bibitem{dai2017contrastive}
B.~Dai and D.~Lin.
\newblock Contrastive learning for image captioning.
\newblock In {\em Advances in Neural Information Processing Systems 30: Annual
  Conference on Neural Information Processing Systems 2017 (NIPS 2017)}, pages
  898--907, 2017.

\bibitem{donahue2015long}
J.~Donahue, L.~Anne~Hendricks, S.~Guadarrama, M.~Rohrbach, S.~Venugopalan,
  K.~Saenko, and T.~Darrell.
\newblock Long-term recurrent convolutional networks for visual recognition and
  description.
\newblock In {\em Proceedings of 2015 IEEE Conference on Computer Vision and
  Pattern Recognition (CVPR 2015)}, pages 2625--2634, 2015.

\bibitem{DongLY15}
C.~Dong, M.~C. Leu, and Z.~Yin.
\newblock American sign language alphabet recognition using {Microsoft}
  {Kinect}.
\newblock In {\em Proceedings of 2015 IEEE Conference on Computer Vision and
  Pattern Recognition (CVPR 2015)}, pages 44--52, 2015.

\bibitem{DuanBSBTCAGOR19}
J.~Duan, G.~Bello, J.~Schlemper, W.~Bai, T.~Dawes, C.~Biffi, A.~de~Marvao,
  G.~Doumou, D.~O'Regan, and D.~Rueckert.
\newblock Automatic {3D} bi-ventricular segmentation of cardiac images by a
  shape-refined multi-task deep learning approach.
\newblock {\em IEEE Transactions on Medical Imaging}, pages 1--1, 2019.

\bibitem{DuanSBDBDDOR18}
J.~Duan, J.~Schlemper, W.~Bai, T.~J.~W. Dawes, G.~Bello, G.~Doumou,
  A.~De~Marvao, D.~P. O'Regan, and D.~Rueckert.
\newblock Deep nested level sets: Fully automated segmentation of cardiac mr
  images in patients with pulmonary hypertension.
\newblock In {\em Proceedings of the 22nd International Conference on Medical
  Image Computing and Computer Assisted Intervention (MICCAI 2018)}, pages
  595--603, 2018.

\bibitem{FangXTL17}
H.~Fang, S.~Xie, Y.~Tai, and C.~Lu.
\newblock {RMPE:} regional multi-person pose estimation.
\newblock In {\em Proceedings of 2017 {IEEE} International Conference on
  Computer Vision (ICCV 2017)}, pages 2353--2362, 2017.

\bibitem{ForsterSHKZPN12}
J.~Forster, C.~Schmidt, T.~Hoyoux, O.~Koller, U.~Zelle, J.~Piater, and H.~Ney.
\newblock Rwth-phoenix-weather: A large vocabulary sign language recognition
  and translation corpus.
\newblock In {\em Proceedings of the 8th International Conference on Language
  Resources and Evaluation (LREC 2012)}, 2012.

\bibitem{ForsterSKBN14}
J.~Forster, C.~Schmidt, O.~Koller, M.~Bellgardt, and H.~Ney.
\newblock Extensions of the sign language recognition and translation corpus
  rwth-phoenix-weather.
\newblock In {\em Proceedings of the 9th International Conference on Language
  Resources and Evaluation (LREC 2014)}, pages 1911--1916, 2014.

\bibitem{gao2017dynamic}
M.~Gao, R.~Yu, A.~Li, V.~I. Morariu, and L.~S. Davis.
\newblock Dynamic zoom-in network for fast object detection in large images.
\newblock In {\em Proceedings of 2017 IEEE Conference on Computer Vision and
  Pattern Recognition (CVPR 2017)}, pages 21--26, 2017.

\bibitem{gattupalli2016evaluation}
S.~Gattupalli, A.~Ghaderi, and V.~Athitsos.
\newblock Evaluation of deep learning based pose estimation for sign language
  recognition.
\newblock In {\em Proceedings of the 9th ACM International Conference on
  PErvasive Technologies Related to Assistive Environments (PETRA 2016)}, pages
  12:1--12:7, 2016.

\bibitem{HeGDG17}
K.~He, G.~Gkioxari, P.~Doll{\'{a}}r, and R.~B. Girshick.
\newblock Mask {R-CNN}.
\newblock In {\em Proceedings of 2017 {IEEE} International Conference on
  Computer Vision (ICCV 2017)}, pages 2980--2988, 2017.

\bibitem{HeZRS16}
K.~He, X.~Zhang, S.~Ren, and J.~Sun.
\newblock Deep residual learning for image recognition.
\newblock In {\em Proceedings of 2016 {IEEE} Conference on Computer Vision and
  Pattern Recognition ({CVPR} 2016)}, pages 770--778, 2016.

\bibitem{HochreiterS97}
S.~Hochreiter and J.~Schmidhuber.
\newblock Long short-term memory.
\newblock {\em Neural Computation}, 9(8):1735--1780, 1997.

\bibitem{HofferHS17}
E.~Hoffer, I.~Hubara, and D.~Soudry.
\newblock Train longer, generalize better: closing the generalization gap in
  large batch training of neural networks.
\newblock In {\em Advances in Neural Information Processing Systems 30: Annual
  Conference on Neural Information Processing Systems (NIPS 2017)}, pages
  1729--1739, 2017.

\bibitem{husqueeze}
J.~Hu, L.~Shen, and G.~Sun.
\newblock Squeeze-and-excitation networks.
\newblock In {\em Proceedings of 2018 IEEE Conference on Computer Vision and
  Pattern Recognition (CVPR 2018)}, pages 7132--7141, 2018.

\bibitem{huang2017densely}
G.~Huang, Z.~Liu, L.~Van Der~Maaten, and K.~Q. Weinberger.
\newblock Densely connected convolutional networks.
\newblock In {\em CVPR}, pages 4700--4708, 2017.

\bibitem{IoffeS15}
S.~Ioffe and C.~Szegedy.
\newblock Batch normalization: Accelerating deep network training by reducing
  internal covariate shift.
\newblock In {\em Proceedings of the 32nd International Conference on Machine
  Learning ({ICML} 2015)}, pages 448--456, 2015.

\bibitem{KarpathyTSLSF14}
A.~Karpathy, G.~Toderici, S.~Shetty, T.~Leung, R.~Sukthankar, and L.~Fei-Fei.
\newblock Large-scale video classification with convolutional neural networks.
\newblock In {\em Proceedings of 2014 IEEE Conference on Computer Vision and
  Pattern Recognition (CVPR 2014)}, pages 1725--1732, 2014.

\bibitem{KimK16}
T.~Kim and S.~Kim.
\newblock Sign language translation system using latent feature values of sign
  language images.
\newblock In {\em Proceedings of the 13th International Conference on
  Ubiquitous Robots and Ambient Intelligence (URAI 2016)}, pages 228--233,
  2016.

\bibitem{KingmaB14}
D.~P. Kingma and J.~Ba.
\newblock Adam: {A} method for stochastic optimization.
\newblock In {\em Proceedings of the 3rd International Conference on Learning
  Representations (ICLR 2015)}, 2015.

\bibitem{KishoreSK14}
P.~V.~V. Kishore, A.~S. C.~S. Sastry, and A.~Kartheek.
\newblock Visual-verbal machine interpreter for sign language recognition under
  versatile video backgrounds.
\newblock In {\em Proceedings of the 1st International Conference on Networks
  Soft Computing (ICNSC 2014)}, pages 135--140, 2014.

\bibitem{KoSJ18}
S.~Ko, J.~G. Son, and H.~D. Jung.
\newblock Sign language recognition with recurrent neural network using human
  keypoint detection.
\newblock In {\em Proceedings of the 2018 Conference on Research in Adaptive
  and Convergent Systems ({RACS} 2018)}, pages 326--328, 2018.

\bibitem{KollerFN15}
O.~Koller, J.~Forster, and H.~Ney.
\newblock Continuous sign language recognition: Towards large vocabulary
  statistical recognition systems handling multiple signers.
\newblock {\em Computer Vision and Image Understanding}, 141:108--125, 2015.

\bibitem{koller2017re}
O.~Koller, S.~Zargaran, and H.~Ney.
\newblock Re-sign: Re-aligned end-to-end sequence modelling with deep recurrent
  cnn-hmms.
\newblock In {\em Proceedings of 2017 IEEE Conference on Computer Vision and
  Pattern Recognition (CVPR 2017)}, pages 3416--3424, 2017.

\bibitem{Liddell03}
S.~K. Liddell.
\newblock {\em Grammar, Gesture, and Meaning in American Sign Language}.
\newblock Cambridge University Press, 2003.

\bibitem{Lin08}
C.-Y. Lin.
\newblock Rouge: A package for automatic evaluation of summaries.
\newblock In {\em Proceedings of ACL workshop on Text Summarization Branches
  Out}, 2004.

\bibitem{liu2017attention}
C.~Liu, J.~Mao, F.~Sha, and A.~L. Yuille.
\newblock Attention correctness in neural image captioning.
\newblock In {\em Proceedings of the 31th {AAAI} Conference on Artificial
  Intelligence (AAAI 2017)}, pages 4176--4182, 2017.

\bibitem{long2015fully}
J.~Long, E.~Shelhamer, and T.~Darrell.
\newblock Fully convolutional networks for semantic segmentation.
\newblock In {\em Proceedings of 2015 IEEE Conference on Computer Vision and
  Pattern Recognition (CVPR 2015)}, pages 3431--3440, 2015.

\bibitem{LuongPM15}
T.~Luong, H.~Pham, and C.~D. Manning.
\newblock Effective approaches to attention-based neural machine translation.
\newblock In {\em Proceedings of the 2015 Conference on Empirical Methods in
  Natural Language Processing (EMNLP 2015)}, pages 1412--1421, 2015.

\bibitem{luvizon20182d}
D.~C. Luvizon, D.~Picard, and H.~Tabia.
\newblock 2d/3d pose estimation and action recognition using multitask deep
  learning.
\newblock In {\em Proceedings of 2018 IEEE Conference on Computer Vision and
  Pattern Recognition (CVPR 2018)}, pages 5137--5146, 2018.

\bibitem{OberwegerWL15}
M.~Oberweger, P.~Wohlhart, and V.~Lepetit.
\newblock Hands deep in deep learning for hand pose estimation.
\newblock In {\em Proceedings of the 24th Computer Vision Winter Workshop (CVWW
  2015)}, pages 1--10, 2015.

\bibitem{PapineniRWZ02}
K.~Papineni, S.~Roukos, T.~Ward, and W.-J. Zhu.
\newblock Bleu: A method for automatic evaluation of machine translation.
\newblock In {\em Proceedings of the 40th Annual Meeting on Association for
  Computational Linguistics (ACL 2002)}, pages 311--318, 2002.

\bibitem{ParkC14}
E.~L. Park and S.~Cho.
\newblock Konlpy: Korean natural language processing in python.
\newblock In {\em Proceedings of the 26th Annual Conference on Human \&
  Cognitive Language Technology (HCLT 2014)}, 2014.

\bibitem{PascanuMB13}
R.~Pascanu, T.~Mikolov, and Y.~Bengio.
\newblock On the difficulty of training recurrent neural networks.
\newblock In {\em Proceedings of the 30th International Conference on Machine
  Learning ({ICML} 2013)}, pages 1310--1318, 2013.

\bibitem{PaszkeGCCYDLDAL17}
A.~Paszke, S.~Gross, S.~Chintala, G.~Chanan, E.~Yang, Z.~DeVito, Z.~Lin,
  A.~Desmaison, L.~Antiga, and A.~Lerer.
\newblock Automatic differentiation in pytorch.
\newblock In {\em NIPS-W}, 2017.

\bibitem{PerezW17}
L.~Perez and J.~Wang.
\newblock The effectiveness of data augmentation in image classification using
  deep learning.
\newblock {\em CoRR}, abs/1712.04621, 2017.

\bibitem{PfisterCZ13}
T.~Pfister, J.~Charles, and A.~Zisserman.
\newblock Large-scale learning of sign language by watching {TV} (using
  co-occurrences).
\newblock In {\em British Machine Vision Conference 2013 (BMVC 2013)}, 2013.

\bibitem{redmon2016you}
J.~Redmon, S.~Divvala, R.~Girshick, and A.~Farhadi.
\newblock You only look once: Unified, real-time object detection.
\newblock In {\em Proceedings of 2016 IEEE Conference on Computer Vision and
  Pattern Recognition (CVPR 2016)}, pages 779--788, 2016.

\bibitem{SimonJMS17}
T.~Simon, H.~Joo, I.~A. Matthews, and Y.~Sheikh.
\newblock Hand keypoint detection in single images using multiview
  bootstrapping.
\newblock In {\em Proceedings of 2017 {IEEE} Conference on Computer Vision and
  Pattern Recognition (CVPR 2017)}, pages 4645--4653, 2017.

\bibitem{SimonyanZ14a}
K.~Simonyan and A.~Zisserman.
\newblock Very deep convolutional networks for large-scale image recognition.
\newblock In {\em Proceedings of the 3rd International Conference on Learning
  Representations (ICLR 2015)}, 2015.

\bibitem{SmithKL17}
S.~L. Smith, P.~Kindermans, C.~Ying, and Q.~V. Le.
\newblock Don't decay the learning rate, increase the batch size.
\newblock In {\em Proceedings of the 6th International Conference on Learning
  Representations (ICLR 2018)}, 2018.

\bibitem{StarnerP95}
T.~Starner and A.~Pentland.
\newblock Real-time {A}merican sign language recognition from video using
  hidden markov models.
\newblock In {\em Proceedings of International Symposium on Computer Vision
  (ISCV 1995)}, pages 265--270, 1995.

\bibitem{SutskeverVL14}
I.~Sutskever, O.~Vinyals, and Q.~V. Le.
\newblock Sequence to sequence learning with neural networks.
\newblock In {\em Advances in Neural Information Processing Systems 27: Annual
  Conference on Neural Information Processing Systems 2014 (NIPS 2014)}, pages
  3104--3112, 2014.

\bibitem{UlyanovVL16}
D.~Ulyanov, A.~Vedaldi, and V.~S. Lempitsky.
\newblock Instance normalization: The missing ingredient for fast stylization.
\newblock {\em CoRR}, abs/1607.08022, 2016.

\bibitem{VaswaniSPUJGKP17}
A.~Vaswani, N.~Shazeer, N.~Parmar, J.~Uszkoreit, L.~Jones, A.~N. Gomez,
  L.~Kaiser, and I.~Polosukhin.
\newblock Attention is all you need.
\newblock In {\em Advances in Neural Information Processing Systems 30: Annual
  Conference on Neural Information Processing Systems 2017 (NIPS 2017)}, pages
  6000--6010, 2017.

\bibitem{VedantamZP15}
R.~Vedantam, C.~L. Zitnick, and D.~Parikh.
\newblock Cider: Consensus-based image description evaluation.
\newblock In {\em Proceedings of 2015 {IEEE} Conference on Computer Vision and
  Pattern Recognition (CVPR 2015)}, pages 4566--4575, 2015.

\bibitem{von2008significance}
U.~Von~Agris, M.~Knorr, and K.-F. Kraiss.
\newblock The significance of facial features for automatic sign language
  recognition.
\newblock In {\em Proceedings of the 8th IEEE International Conference on
  Automatic Face and Gesture Recognition (FG 2008)}, pages 1--6, 2008.

\bibitem{WeiRKS16}
S.~Wei, V.~Ramakrishna, T.~Kanade, and Y.~Sheikh.
\newblock Convolutional pose machines.
\newblock In {\em Proceedings of 2016 {IEEE} Conference on Computer Vision and
  Pattern Recognition (CVPR 2016)}, pages 4724--4732, 2016.

\bibitem{WuJ19}
Y.~Wu and Q.~Ji.
\newblock Facial landmark detection: A literature survey.
\newblock {\em International Journal of Computer Vision}, 127(2):115--142,
  2019.

\bibitem{xu2015show}
K.~Xu, J.~Ba, R.~Kiros, K.~Cho, A.~Courville, R.~Salakhudinov, R.~Zemel, and
  Y.~Bengio.
\newblock Show, attend and tell: Neural image caption generation with visual
  attention.
\newblock In {\em Proceedings of the 32nd International Conference on Machine
  Learning ({ICML} 2015)}, pages 2048--2057, 2015.

\bibitem{zhang2018context}
H.~Zhang, K.~Dana, J.~Shi, Z.~Zhang, X.~Wang, A.~Tyagi, and A.~Agrawal.
\newblock Context encoding for semantic segmentation.
\newblock In {\em Proceedings of 2018 IEEE Conference on Computer Vision and
  Pattern Recognition (CVPR 2018)}, pages 7151--7160, 2018.

\end{thebibliography}
}

\end{document}